\pdfoutput=1

\documentclass[11pt]{article}

\usepackage[final]{acl}

\usepackage{times}
\usepackage{latexsym}

\usepackage[T1]{fontenc}

\usepackage[utf8]{inputenc}

\usepackage{microtype}

\usepackage{inconsolata}

\usepackage{graphicx}

\usepackage{amsmath}
\usepackage{multicol}
\usepackage{times}
\usepackage{dsfont}
\usepackage{enumitem}
\usepackage{placeins}
\usepackage{colortbl}

\definecolor{level7}{rgb}{1.0, 1.0, 1.0}
\definecolor{level6}{rgb}{0.96, 0.96, 0.96}
\definecolor{level5}{rgb}{0.92, 0.92, 0.92}
\definecolor{level4}{rgb}{0.88, 0.88, 0.88}
\definecolor{level3}{rgb}{0.84, 0.84, 0.84}
\definecolor{level2}{rgb}{0.80, 0.80, 0.80}
\definecolor{level1}{rgb}{0.76, 0.76, 0.76}
\definecolor{level0}{rgb}{0.72, 0.72, 0.72}

\newcommand{\mycomment}[1]{}

\title{Measuring Social Biases in Masked Language Models by Proxy of Prediction Quality}

\author{%
  Rahul Zalkikar\thanks{*While collaboration between Zalkikar and Chandra inspired this work, the core methods proposed herein are attributable to Zalkikar, the corresponding author.} \\
  New York University \\
  \texttt{rz1567@nyu.edu}
    \\\And
   Kanchan Chandra\footnotemark[1] \\
    New York University \\
    \texttt{kc76@nyu.edu}
}

\begin{document}

\maketitle

\begin{abstract}

Transformer language models have achieved state-of-the-art performance for a variety of natural language tasks but have been shown to encode unwanted biases. We evaluate the social biases encoded by transformers trained with the masked language modeling objective using proposed proxy functions within an iterative masking experiment to measure the quality of transformer models’ predictions and assess the preference of MLMs towards disadvantaged and advantaged groups. We find all models encode concerning social biases. We compare bias estimations with those produced by other evaluation methods using benchmark datasets and assess their alignment with human annotated biases. We extend previous work by evaluating social biases introduced after retraining an MLM under the masked language modeling objective and find proposed measures produce more accurate and sensitive estimations of biases introduced by retraining MLMs based on relative preference for biased sentences between models, while other methods tend to underestimate biases after retraining on sentences biased towards disadvantaged groups.

\small\textit{\textbf{Warning:} This paper contains explicit statements of biased stereotypes and may be upsetting.}
\end{abstract}

\section{Introduction}

Masked language models (MLM(s); \citealp{brown2020language,devlin-etal-2019-bert,liu2019improving,radford2019language}) such as transformers BERT and RoBERTa incorporate bidirectionality and self-attention, producing contextually-aware embeddings. Unlike $\text{BERT}_{\text{unc}}$, RoBERTa was pretrained solely on the masked language modeling objective (MLMO) on a larger corpus of text. RoBERTa uses a dynamic masking strategy for a diverse set of representations during training and has achieved state-of-the-art results on GLUE, RACE, and SQuAD (\citealp{DBLP:journals/corr/abs-1907-11692,rajpurkar-etal-2016-squad,lai-etal-2017-race,wang-etal-2018-glue}). Distilled model variants have been shown to train significantly faster for minor decreases in performance \cite{DBLP:journals/corr/abs-1910-01108}. MLMs have produced state-of-the-art results for masked language modeling, named entity recognition, and intent or topic classification, but also encode concerning social biases against disadvantaged groups that are undesirable in production settings. As MLMs become increasingly prevalent, researchers have been working on methods to measure biases embedded in these models \cite{nangia-etal-2020-crows,Kaneko_Bollegala_2022,10.1007/978-3-031-33374-3_42}.

To address the issue of social biases in MLMs at its source, we measure biases of MLMs while focusing on an MLM's key pretraining objective: masked language modeling. In this work, we focus on proposing and assessing bias evaluation measures for MLMs and \textit{not} proposing methods for debiasing MLMs. However, when assessing evaluation measures, we consider prior research that involves retraining or fine-tuning under the MLMO with debiased or counterfactual data to reduce biases in MLMs, such as \citealp{zhao-etal-2018-gender} and \citealp{zhao-etal-2019-gender} which use data augmentation to swap gendered words with their opposites prior to retraining. Motivated by this, we assess whether proposed measures satisfy an important criterion for improvement over previously proposed ones: alignment with biases introduced by MLM retraining under the MLMO (Section \ref{awrb}).

We represent MLM bias through a model’s relative preference for ground truth tokens between two paired biased sentences, where one contains bias against disadvantaged groups (stereotypical) and the other contains bias against advantaged groups (anti-stereotypical). We assess the relative preferences of MLMs towards sentences biased against disadvantaged and advantaged groups using proxy measures for prediction quality. In particular, \textbf{we quantify MLM preference by proxy of masked token prediction quality given unmasked token context between encoded sentences within pairs}. 

To measure the preference of an MLM using the (attention-weighted) quality of its predictions under the MLMO, we propose and validate proxy functions $\Delta\textsc{pa}$ (Equation \ref{dpa}; \textbf{P}robability Difference with \textbf{A}ttention) and \textsc{crra} (Equation \ref{crra}; \textbf{C}omplementary \textbf{R}eciprocal \textbf{R}ank with \textbf{A}ttention) and introduce a modified $\Delta\textsc{p}$ (Equation \ref{dp-alt}) to measure the likelihood an MLM will select a ground truth token to replace a masked one, and extend these definitions for a sentence. We apply per-model indicator function \textsc{bspt} (Equation \ref{dmse1}) to estimate the encoded social biases in pretrained MLMs. Our approach differs from prior research in measuring biases in MLMs by using attention-weights under an \textbf{I}terative \textbf{M}asking \textbf{E}xperiment (\textbf{\textsc{ime}}; Section \ref{experiments}) to probe MLM preferences. 

We compare pre- and retrained MLMs within the same model class to recover the nature of biases introduced by retraining. In particular, we define and apply a proxy for the relative preference between two MLMs with model-comparative indicator function \textbf{\textsc{bsrt}} (\textbf{B}ias \textbf{S}core for MLM \textbf{Re}-\textbf{t}raining; Equation \ref{dmse2}) to estimate social biases introduced by retraining MLMs under the MLMO.\footnote{We acknowledge that "introduced" biases must be represented by the relative change in biases after retraining an MLM under the MLMO.}

In summary, the primary contributions of this work are as follows:
\begin{itemize}[topsep=0em, leftmargin=4mm]
    \setlength{\itemsep}{0pt}
    \setlength{\parskip}{0pt}
    \setlength{\parsep}{0pt}
    \item We explore MLM bias through a model’s relative preference for ground truth tokens between two minimally distant sentences with contrasting social bias under the \textsc{ime}. We measure this using the attention-weighted quality of predictions.
    \item We propose model-comparative indicator function \textsc{bsrt} to estimate social biases against disadvantaged groups for a retrained MLM relative to its pretrained base, and assess bias evaluation measures for alignment with biases introduced by MLM retraining under the MLMO.
    \item We evaluate social biases for four transformer models available through the Transformers library \cite{wolf-etal-2020-transformers}. We use proxy measures for MLM prediction quality with model-comparative function \textsc{bsrt} to estimate social biases introduced by retraining MLMs under the MLMO. We find proposed measures $\Delta\textsc{pa}$ and \textsc{crra}, along with modified $\Delta\textsc{p}$ and \textsc{crr}, produce more accurate estimations of biases introduced by MLM retraining than previously proposed ones, which can produce concerning underestimations of biases after retraining MLMs on sentences biased against disadvantaged groups. We observe that proposed measures $\Delta\textsc{pa}$ and \textsc{crra}, along with modified $\Delta\textsc{p}$, show greater sensitivity than \textsc{crr}, \textsc{csps}, \textsc{aul}, and \textsc{aula} to relative changes in MLM bias due to retraining, indicated by larger and smaller bias scores and more frequently significant relative difference in proportions of bias between pre- and retrained transformers.
\end{itemize}

Our methodology could help others evaluate social biases encoded in an MLM after it is retrained on the MLMO, such as for any downstream fill-mask task, or after it is fine-tuned for other objectives that alter weights and change MLMO performance. We release a package for measuring biases in MLMs to enable bias score computation on user-supplied or benchmark datasets and easy integration into existing evaluation pipelines.

\section{Related Work} \label{relatedwork}

\subsection{Biases in Static Word Embeddings}

Static word embeddings \cite{pennington-etal-2014-glove,mikolov2013efficient} can be shifted in a direction to decompose bias embedded in learned text data representations. These could be analogies or biases along an axis, such as gender (\citealp{NIPS2016_a486cd07}) or race (\citealp{manzini-etal-2019-black}). The Word Embedding Association Test (WEAT; \citealp{doi:10.1126/science.aal4230}) measures association between targets and attributes using cosine similarity between static word embeddings, but has been shown to overestimate biases by \citealp{ethayarajh-etal-2019-understanding} which proposed the robust relational inner product association (RIPA) method, derived from the subspace projection method to debias vectors in \citealp{NIPS2016_a486cd07}.

While WEAT has shown that token-level embeddings produced by GloVe and Word2Vec encode biases based on gender and race \cite{doi:10.1126/science.aal4230}, the Sentence Encoder Association Test (SEAT; \citealp{may-etal-2019-measuring}) extends on WEAT to measure social biases in sentence-level encoders such as ELMo and BERT using template sentences with masked target tokens, averaging token embeddings to form sentence-level embeddings on which cosine similarity is applied as a proxy for semantic association. As an alternative evaluation method under a different objective, \citealp{liang-etal-2020-monolingual} assessed differences in log-likelihood between gender pronouns in a template sentence where occupations can uncover the directionality of the bias encoded by an MLM.

\subsection{Evaluating Biases in MLMs} \label{relatedwork-biaseval}

When considering a sentence $s$ containing tokens $\{t_{1}, t_{2}, ..., t_{l_{s}}\}$, where $l_{s}$ is the number of tokens in $s$, (modified) token(s) of $s$ can characterize its bias towards either disadvantaged or advantaged groups. For a given sentence $s$ with tokens $t\in s$ we denote all tokens besides $t_{x}$ as $s_{\setminus t_{x}}$ (where $1 \leq x \leq l_{s}$), and we denote modified tokens as $M$ and unmodified tokens as $U$ ($s = U \cup M$). For a given MLM with parameters $\theta$, we denote a masked token as $t_{m}$ and a predicted token as $t_{p}$.

\citealp{salazar-etal-2020-masked} uses pseudo-log-likelihood scoring to approximate $P(U|M, \theta)$, or the probability of unmodified tokens conditioned on modified ones. Similarly, \citealp{nangia-etal-2020-crows} reports CrowS-Pairs Scores (\textsc{csps}; Appendix \ref{appendix:d}), a pseudo-log-likelihood score for an MLM selecting unmodified tokens given modified ones. \citealp{nadeem-etal-2021-stereoset} reports a StereoSet Score (\textsc{sss}; Appendix \ref{appendix:e}), a pseudo-log-likelihood score for an MLM selecting modified tokens given unmodified ones. 

\citealp{10.1007/978-3-031-33374-3_42} tests MLMs in an iterative fill mask setting where the model outputs a set of tokens (or the (log)softmax of model logits mapped to tokens) to fill the masked one, starting with the token of highest probability $P(t_{p}|c)$ and, as such, first rank $\rho(t_{p}|c) = 1$ in the set of model token predictions (limited by the MLM's embedding space).

{
\small
\begin{equation} \label{dp}
\begin{split}
\Delta \textsc{p}(t|s_{\setminus t_{m}}; \theta) &= P(t_p|s_{\setminus t_{m}}; \theta) - P(t_m|s_{\setminus t_{m}}; \theta) \\
&= \Delta \textsc{p}(w; \theta)
\end{split}
\end{equation}
}

$\Delta \textsc{p}(t|s_{\setminus t_{m}}; \theta)$ (Equation \ref{dp}) represents the difference between the probability of a predicted token $t_{p}$ and a masked token $t_{m}$ in a sentence $s$. It serves as a proxy of the MLM's prediction quality for a token given its context within the \textsc{ime}, or all tokens in $s$ besides $t_{m}$ (\citealp{10.1007/978-3-031-33374-3_42}). 

{
\small
\begin{equation} \label{crr}
\begin{split}
\textsc{crr}(t|s_{\setminus t_{m}}; \theta) &= \bigl(\rho(t_{p}|s_{\setminus t_{m}}; \theta)^{-1} - \rho(t_{m}|s_{\setminus t_{m}}; \theta)^{-1}\bigr) \\
&= 1 -\rho(t_{m}|s_{\setminus t_{m}}; \theta)^{-1} \\
&= \textsc{crr}(w; \theta)
\end{split}
\end{equation}
}

\citealp{10.1007/978-3-031-33374-3_42} proposes the Complementary Reciprocal Rank ($\textsc{crr}(t|s_{\setminus t_{m}}; \theta)$; Equation \ref{crr}) for a masked token given its context, where $\rho(t_{p}|s_{\setminus t_{m}})^{-1}$ is the reciprocal rank of the predicted token (and always equal to 1) and $\rho(t_{m}|s_{\setminus t_{m}})^{-1}$ is the reciprocal rank of the masked token. Thus, $\rho(t_{m}|s_{\setminus t_{m}})^{-1}$ provides a likelihood measure for $t_{m}$ being chosen by the model as a candidate token to replace the ground truth (masked) one. 

\citealp{10.1007/978-3-031-33374-3_42} defines $\Delta \textsc{p}(s)$ (Appendix \ref{appendix:h}) as the probability difference for a sentence $s$ and $\textsc{crr}(s)$ (Appendix \ref{appendix:i}; average of all single masked token \textsc{crr}s with token ordering preserved) as the complementary reciprocal rank for a sentence $s$, and uses them as proxies for an MLM's prediction quality or preference.%

\citealp{Kaneko_Bollegala_2022} propose evaluation metrics All Unmasked Likelihood (\textsc{aul}; Appendix \ref{appendix:j}) and \textsc{aul} with Attention weights (\textsc{aula}; Appendix \ref{appendix:k}), where both are generated by requiring the MLM to predict all tokens (unmasked input). By requiring the MLM to simultaneously predict all of the tokens in a given unmasked input sentence $s$, \citealp{Kaneko_Bollegala_2022} aim to eliminate biases associated with masked tokens under previously proposed pseudo-likelihood-based scoring methods (\citealp{nadeem-etal-2021-stereoset}, \citealp{nangia-etal-2020-crows}), which assumed that masked tokens are statistically independent, and selectional biases from masking a subset of input tokens, such as high frequency words (masked more often during training).

\citealp{Kaneko_Bollegala_2022} reference the use of sentence-level embeddings produced by MLMs for downstream tasks such as sentiment classification to argue that biases associated with masked tokens should not influence the intrinsic bias evaluation of an MLM, as opposed to the evaluation of biases introduced after an MLM is fine-tuned. In contrast, we focus on an MLM's key pretraining objective, masked language modeling, to measure social biases of the MLM.\footnote{Masked language modeling was a pretraining objective for all transformers considered in this work. Next sentence prediction was not used for RoBERTa and variants due to relatively lower performance with its inclusion \cite{DBLP:journals/corr/abs-1907-11692}.} In addition, we measure relative changes in biases w.r.t. the intrinsic biases of the same base MLM after retraining under the MLMO (as opposed to fine-tuning). Thus, we argue that biases associated with masked tokens are not undesirable in our case.

\textsc{aul} and \textsc{aula} were found to be sensitive to contextually meaningful inputs by randomly shuffling tokens in input sentences and comparing accuracies with and without shuffle. \textsc{crr} is also conditional to the unmasked token context by definition. We argue measure sensitivity to unmasked token contexts is desirable when evaluating a given MLM's preference under a fill-mask task, or when estimating biases using contextualized token-level embeddings produced by an MLM. 

In contrast with previous methods such as \textsc{sss} and \textsc{csps}, we refrain from using strictly modified or unmodified subsets of input tokens as context and instead provide all tokens but the ground truth one as context for MLM prediction under each iteration of our masking experiment. In this sense, \textsc{crr} and $\Delta \textsc{p}$ consider all tokens equally, and like \textsc{aul}, they might also benefit from considering the weight of MLM attention as a proxy for token importance when probing for MLM preferences for ground truth tokens between two paired sentences along a social bias axis.\footnote{By incorporating attention weights, $\Delta\textsc{pa}$ and \textsc{crra} consider token-level contributions to MLM predictions, leveraging the attention mechanism’s role in weighting contextually important tokens.}

Existing benchmark datasets such as CPS are limited to one ground truth per masked token, so an important consideration is an MLM's ability to predict multiple plausible tokens for a context that could qualify for concerning social bias but goes unrecognized during evaluation using previously proposed measures. \textsc{crr} could perform better than pseudo-(log)likelihood-based measures for this sensitivity that yields larger relative differences in $\Delta \text{P}(t|c)$ as opposed to $\textsc{crr}(t|c)$\footnote{The lowest rank of any possible ground truth token within a model's vocabulary is equal to the size of the vocabulary ($1 \leq \rho(t_{m}|c) \leq V$ and $0 \leq \textsc{crr}(s) \leq 1-(V)^{-1}$ for an MLM with embedding vocabulary of size $V$).}, and is deemed critical for evaluation by \citealp{10.1007/978-3-031-33374-3_42} due to the possible uniformity of probabilities generated by a particular MLM w.r.t. others.

\section{Methodology} \label{methodology}
\subsection{Biases in Pretrained MLMs} \label{prevspre}

\begin{table*}[htb]
\small\centering
\begin{tabular}{l|l|l}
\hline
\textbf{Model} & \textbf{\textsc{ime} Input} with masked token $w_{11}$ = money & \textbf{$w_{11}$ Rank} \\
\hline
$\text{distilRoBERTa}_{P}$ & The \textit{poor} are really ignorant about how to handle the \textbf{problems} they do have & \textbf{5} \\
$\text{distilRoBERTa}_{R}$ & The \textit{poor} are really ignorant about how to handle the \textbf{money} they do have & \textbf{1} \\
$\text{distilRoBERTa}_{P}$ & The \textit{rich} are really ignorant about how to handle the \textbf{wealth} they do have & \textbf{4} \\
$\text{distilRoBERTa}_{R}$ & The \textit{rich} are really ignorant about how to handle the \textbf{problems} they do have & \textbf{2} \\
\hline
\end{tabular}
\caption{MLM predictions and ground truth masked token ranks for input sentences under the \textsc{ime}. The processed input sentences are "The [\textit{poor}, \textit{rich}] are really ignorant about how to handle the money they do have", and are contained in $S_{\text{dis}}$ or $S_{\text{adv}}$ from CPS with characteristic words \textit{poor} and \textit{rich} respectively. Masked tokens are in bold.}
\label{mlmpred-main}
\end{table*}

We probe for MLM preferences using the \textsc{ime}, which masks one token at a time until all tokens have been masked, or until we have $n$ logits or predictions for a sentence with $n$ tokens. Appendix \ref{appendix:l} shows an \textsc{ime} example for one model and text. 

Table \ref{mlmpred-main} shows masked token predictions (those with first rank and highest probability) produced by MLMs for example input contexts. The ground truth masked token rank, or $\rho(t_{m}|c)$, represents the quality of the models' predictions relative to a sentence biased against an advantaged or disadvantaged group. Special start and end character tokens for MLMs are not included in the span of tokens considered in our experiments to eliminate noise.\footnote{Special start and end character tokens for MLMs are not considered by measures using attention weights and probabilities computed from the (log)softmax of model logits for the masked token index.}

{
\small
\begin{equation} \label{dp-alt}
\begin{split}
\Delta \text{P}(t|s_{\setminus t_{m}}; \theta) 
&= \bigl(\log P(t_{p}|s_{\setminus t_{m}}; \theta) - \log P(t_{m}|s_{\setminus t_{m}}; \theta)\bigr) \\
&= \Delta \textsc{P}(w; \theta)
\end{split}
\end{equation}
}

Motivated by the typically long-tail distribution of probabilities output by MLMs, we redefine Equation \ref{dp} and propose a modified version $\Delta \text{P}(w)$ as shown in Equation \ref{dp-alt}. The log transformation reduces sensitivity to differences when probabilities are high but enhances relative differences when probabilities are low, improving sensitivity to subtle but potentially meaningful variations in low-probability token choices.

{
\small
\begin{equation} \label{crra-tc}
\begin{split}
\textsc{crra}(t|s_{\setminus t_{m}}; \theta) &= a_{m} \bigl(1 - \log\rho(t_{m}|s_{\setminus t_{m}}; \theta)^{-1}\bigr) \\
&= \textsc{crra}(w;\theta)
\end{split}
\end{equation}
}
{
\small
\begin{equation} \label{pa-tc}
\begin{split}
\Delta\textsc{pa}(t|s_{\setminus t_{m}}; \theta) &= a_{m}\bigl(\log P(t_{p}|s_{\setminus t_{m}}; \theta)\\
&\hspace{32pt} -\log P(t_{m}|s_{\setminus t_{m}}; \theta)\bigr)\\
&= \Delta\textsc{pa}(w;\theta)
\end{split}
\end{equation}
}

For a given MLM with parameters $\theta$, we propose attention-weighted measures $\textsc{crra}(w)$ and $\Delta \text{PA}(w)$ defined in Equations \ref{crra-tc} and \ref{pa-tc} respectively, where $a_{w}$ is the average of all multi-head attentions associated with the ground truth token $w$\footnote{Averaging the multi-head attentions associated with a ground truth token provides a potentially balanced representation of token-level contributions by capturing diverse contextual relationships encoded by individual heads while reducing noise from any single head.}, and $P(t_{m}|s_{\setminus t_{m}})$ and $\rho(t_{m}|s_{\setminus t_{m}})$ are the probability score and rank of the masked token respectively. We extend these definitions for a sentence $s$ as shown in Equations \ref{crra} and \ref{dpa}.

{
\small
\begin{equation} \label{crra} 
\textsc{crra}(s) := \frac{1}{l_{s}} \sum_{w \in s} \textsc{crra}(w;\theta)
\end{equation}
}
{
\small
\begin{equation} \label{dpa} 
\Delta\textsc{pa}(s) := \frac{1}{l_{s}} \sum_{w \in s} \Delta\textsc{pa}(w;\theta)
\end{equation} 
}

We compute measure $f$, $\forall f\in\{\textsc{crr}_{T}(s),\\\Delta\textsc{p}_{T}(s),\textsc{crra}_{T}(s),\Delta\textsc{pa}_{T}(s)\}$, where $T$ is an MLM transformer and $s$ is a sentence, $\forall s \in S_{\text{dis}}$ and $\forall s \in S_{\text{adv}}$.\footnote{We apply the Shapiro-Wilk test \cite{eb32428d-e089-3d0c-8541-5f3e8f273532} for normality to each measure and did not find evidence that the measures were not drawn from a normal distribution. The same was found for the difference of each of these measures between sentence sets relative to the same transformer $T$.} Appendix \ref{appendix:r} shows these measures (likelihood scores) for an example sentence $s$ in SS and CPS.\footnote{Greater MLM preference based on prediction quality is reflected by \textsc{crr}, \textsc{crra}, $\Delta\textsc{p}$ and $\Delta\textsc{pa}$ values closer to 0 (if a sentence with bias against \textit{advantaged} groups has a \textit{greater} value relative to its paired counterpart, the MLM is deemed to prefer bias against \textit{disadvantaged} groups). The opposite is true for measures \textsc{csps}, \textsc{sss}, \textsc{aul} and \textsc{aula}.} We define sets of measures $M_{1}$ and $M_{2}$ ($M_{1} \cap M_{2} =	\emptyset$), where $M_{1}=$ $\{\textsc{crr},$ $\textsc{crra},$ $\Delta\textsc{pa},$ $\Delta\textsc{p}\}$ and $M_{2}=$ $\{\textsc{aul},$ $\textsc{aula},$ $\textsc{csps},$ $\textsc{sss}\}$.

{
\small
\begin{equation} \label{fdiff}
    \Delta f_{T}(i) = 
\begin{cases}
    f_{T}(S_{\text{adv}}(i)) - f_{T}(S_{\text{dis}}(i)),& \text{if } f \in M_{1} \\
    f_{T}(S_{\text{dis}}(i)) - f_{T}(S_{\text{adv}}(i)),& \text{if } f \in M_{2}
\end{cases}
\end{equation}
}

We apply Equation \ref{fdiff} to estimate the preference of a transformer $T$ for $s \in S_{\text{dis}}$ relative to $s \in S_{\text{adv}}$ for paired sentence $s$ with index $i$ and measure $f$, $\forall f \in$ $\{\textsc{crr},$ $\textsc{crra},$ $\Delta\textsc{p},$ $\Delta\textsc{pa},$ $\text{\textsc{csps}},$ $\textsc{sss},$ $\text{\textsc{aul}},$ $\text{\textsc{aula}}\}$. We define a bias score for a pretrained MLM as \textsc{bspt} (Equation \ref{dmse1}).

{
\small
\begin{equation} \label{dmse1} 
\textsc{bspt}(f) := \frac{100}{N}\sum_{i=1}^{N} \mathds{1}\bigl(\Delta f_{T}(i) > 0\bigr)
\end{equation}
}

$\mathds{1}$ is a per-model indicator function which returns $1$ if transformer $T$ has a larger preference for $S_{\text{dis}}$ relative to $S_{\text{adv}}$ and $0$ otherwise, as estimated for a measure $f$ by Equation \ref{fdiff}. \textsc{bspt} represents the proportion of sentences with higher relative bias against disadvantaged groups for a given measure, where values above 50 indicate greater relative bias against disadvantaged groups for an MLM.

\subsection{Biases Introduced by MLM Retraining} \label{prevsre}

We define a bias score for a retrained MLM (relative to its pretrained base) as \textsc{bsrt} (Equation \ref{dmse2}).

{
\small
\begin{equation} \label{dmse2} 
    \textsc{bsrt}(f) := \frac{100}{N}\sum_{i=1}^{N} \mathds{1}\bigl(\Delta f_{T_{1}}(i) > \Delta f_{T_{2}}(i)\bigr)
\end{equation}
}

We define a proxy for the relative preference between two MLMs with model-comparative indicator function $\mathds{1}$, which returns $1$ if transformer $T_{1}$ has a larger preference for $S_{\text{dis}}$ relative to $S_{\text{adv}}$ than transformer $T_{2}$ and $0$ otherwise, as estimated for a measure $f$ by Equation \ref{fdiff}. \textsc{bsrt} can be applied to compare pre- and retrained MLMs within the same model class and recover biases introduced by MLM retraining. Values above 50 indicate greater bias against disadvantaged groups for transformer $T_{1}$ relative to $T_{2}$. %

\section{Experiments and Findings} \label{experiments}
\subsection{Benchmark Datasets for Social Bias}

The Crowdsourced Stereotype Pairs Benchmark (CPS; \citealp{nangia-etal-2020-crows}) contains biased sentences towards historically advantaged and disadvantaged groups along nine forms of social biases. StereoSet (SS; \citealp{nadeem-etal-2021-stereoset}) contains intrasentence and intersentence (with context) pairs for four forms of social biases, using the likelihood of modified tokens given unmodified token contexts as proxy for MLM preference. Similarly, CPS contains characteristic words that distinguish sentences within pairs and define the nature of a particular bias towards either advantaged or disadvantaged groups, but instead uses the relative likelihood of unmodified tokens being chosen by the MLM given a modified context (characteristic word) across a sentence pair (see Appendix \ref{appendix:a} for details on bias categories and sentence counts). 

To probe for biases of interest that are encoded in MLMs, the scope of our experiments include all bias categories and sentences pairs in CPS and intrasentence pairs in SS, since intersentence pairs are not masked for bias evaluation (\citealp{Kaneko_Bollegala_2022}). We estimate MLM preference towards a stereotypical sentence over a less stereotypical one for each bias category in CPS and SS.

\subsection{Retraining Dataset}

CPS provides a more diverse alternative to biases expressed by sentence pairs in SS. Biases widely acknowledged in the United States are well represented in CPS\footnote{CPS categories are a "narrowed" version of the US Equal Employment Opportunities Commission’s list of protected categories \cite{nangia-etal-2020-crows}.} compared to SS, and there is greater diversity of sentence structures in CPS \cite{nangia-etal-2020-crows}. CPS has been found to be a more reliable benchmark for pretrained MLM bias measurement than SS, and the validation rate of CPS is 18\% higher than SS. \cite{nangia-etal-2020-crows}. Based on these findings, paired with (1) the computational expense and time-consumption involved with retraining MLMs under the MLMO and (2) concerns regarding standard masked language modeling metric viability on SS, we proceed to use sentence sets in CPS to re-train MLMs and validate methods for estimating the biases that are introduced \cite{nangia-etal-2020-crows} (see Appendix \ref{appendix:a.2} for details). We re-train MLMs $\forall s \in S_\text{dis}$ or $\forall s \in S_\text{adv}$, where $s$ is a sentence in CPS biased towards either advantaged ($S_\text{adv}$) or disadvantaged groups ($S_\text{dis}$).%

\subsection{Transformer-based Language Models} \label{models-used}

We denote pre- and retrained transformers as $T_{P}$ and $T_{R}$ respectively. The subscript $_{\text{unc}}$ denotes an uncased model. We report results from the following transformer-based language models available through the HuggingFace library \cite{wolf-etal-2020-transformers}: \textbf{bert-base-uncased} ($\text{BERT}_{\text{unc}}$; \citealp{devlin-etal-2019-bert}), \textbf{roberta-base} (RoBERTa; \citealp{DBLP:journals/corr/abs-1907-11692}), \textbf{distilbert-base-uncased} ($\text{distilBERT}_{\text{unc}}$; \citealp{DBLP:journals/corr/abs-1910-01108}), and \textbf{distilroberta-base} (distilRoBERTa; \citealp{liu2019improving}). 

\subsection{Pretrained MLM Bias Scores} \label{pre-mlm-bias-scores}

We use \textsc{bspt} to compare MLM preferences, report overall bias scores in Table \ref{overallscores}, and provide detailed results for all measures and MLMs in Appendix \ref{appendix:n}.

{
\renewcommand{\arraystretch}{0.96}
\begin{table}[hbt]
\small\centering
\tabcolsep=0.075cm
\scalebox{0.95}{
\begin{tabular}{lcccc}
\hline
\hline
$f$ & $\text{RoBERTa}_{P}$ & $\text{BERT}_{P,\text{unc}}$ & $\text{D-RoBERTa}_{P}$ & $\text{D-BERT}_{P,\text{unc}}$ \\
\hline
\multicolumn{5}{l}{\textbf{CPS Dataset}} \\
\hline
\textbf{\textsc{csps}} & \cellcolor{level3} 59.35 & \cellcolor{level2} 60.48 & \cellcolor{level3} 59.35 & \cellcolor{level3} 56.83 \\
\hline
\textbf{\textsc{aul}} & \cellcolor{level5} 58.75 & \cellcolor{level6} 48.34 & \cellcolor{level5} 53.32 & \cellcolor{level6} 51.59 \\ 
\textbf{\textsc{aula}} & \cellcolor{level6} 58.09 & \cellcolor{level5} 48.21 & \cellcolor{level6} 51.86 & \cellcolor{level5} 52.65 \\ 
\hline
\textbf{\textsc{crr}} & \cellcolor{level4} 58.89 & \cellcolor{level0} \textbf{61.07} & \cellcolor{level4} 57.76 & \cellcolor{level4} 56.23 \\
\textbf{\textsc{crra}} & \cellcolor{level0} \textbf{60.68} & \cellcolor{level4} 58.89 & \cellcolor{level0} \textbf{61.94} & \cellcolor{level0} \textbf{60.08} \\
\textbf{$\Delta\textsc{p}$} & \cellcolor{level2} 59.88 & \cellcolor{level3} 60.08 & \cellcolor{level2} 59.75 & \cellcolor{level2} 57.49 \\
\textbf{$\Delta\textsc{pa}$} & \cellcolor{level1} 60.15 & \cellcolor{level1} 60.81 & \cellcolor{level1} 59.81 & \cellcolor{level1} 58.02 \\
\hline
\multicolumn{5}{l}{\textbf{SS Dataset}} \\
\hline
\textbf{\textsc{sss}} & \cellcolor{level3} 61.06 & \cellcolor{level0} \textbf{59.16} & \cellcolor{level0} \textbf{61.4} & \cellcolor{level1} 60.59 \\
\hline
\textbf{\textsc{aul}} & \cellcolor{level4} 59.45 & \cellcolor{level6} 48.91 & \cellcolor{level3} 60.21 & \cellcolor{level4} 51.71 \\
\textbf{\textsc{aula}} & \cellcolor{level5} 58.83 & \cellcolor{level5} 50.28 & \cellcolor{level4} 59.59 & \cellcolor{level5} 51.66 \\
\hline
\textbf{\textsc{crr}} & \cellcolor{level6} 57.83 & \cellcolor{level4} 53.85 & \cellcolor{level5} 54.37 & \cellcolor{level3} 53.42 \\
\textbf{\textsc{crra}} & \cellcolor{level2} 62.06 & \cellcolor{level2} 58.59 & \cellcolor{level2} 60.54 & \cellcolor{level0} \textbf{61.11} \\
\textbf{$\Delta\textsc{p}$} & \cellcolor{level1} 62.2 & \cellcolor{level1} 58.64 & \cellcolor{level0} \textbf{61.4} & \cellcolor{level2} 59.31 \\
\textbf{$\Delta\textsc{pa}$} & \cellcolor{level0} \textbf{62.35} & \cellcolor{level3} 58.21 & \cellcolor{level1} 61.35 & \cellcolor{level2} 59.31 \\
\hline
\hline
\end{tabular}
}
\caption{Overall bias scores for pretrained MLMs using \textsc{bspt}$(f)$ with measures $\forall f \in$ $\{\textsc{crr},$ $\textsc{crra},$ $\Delta\textsc{p},$ $\Delta\textsc{pa},$ $\text{\textsc{csps}},$ $\textsc{sss},$ $\text{\textsc{aul}},$ $\text{\textsc{aula}}\}$ on CPS and SS, where D- denotes distilled. Color-coded from lightest to darkest, with lower values represented by lighter shades and higher values by darker shades. Bold values indicate the highest bias score for an MLM across all measures.}
\label{overallscores}
\end{table}
}

Overall, all evaluation methods show concerning social biases against disadvantaged groups embedded in MLMs as observed in prior research (\citealp{Kaneko_Bollegala_2022}, \citealp{nangia-etal-2020-crows}). Interestingly, $\text{BERT}_{\text{unc}}$ has the lowest overall \textsc{sss}, \textsc{aul}, \textsc{aula}, \textsc{crra}, $\Delta\textsc{p}$, and $\Delta\textsc{pa}$ (second lowest \textsc{crr}) on SS, but conflicting results on CPS, where it has the highest \textsc{csps}, \textsc{crr}, $\Delta\textsc{p}$, and $\Delta\textsc{pa}$ but the lowest \textsc{aul}, \textsc{aula}, and \textsc{crra}. RoBERTa and distilRoBERTa have higher overall bias than $\text{BERT}_{\text{unc}}$ and $\text{distilBERT}_{\text{unc}}$ according to (1) all but one measure on SS and (2) \textsc{aul} and \textsc{crra} on CPS. 

\citealp{Kaneko_Bollegala_2022} observe a higher bias score for religion in CPS across \textsc{csps}, \textsc{aul}, and \textsc{aula} with the \textbf{roberta-large} MLM. \citealp{nangia-etal-2020-crows} also observe that \textbf{roberta-large} has relatively higher bias scores for the religion category in CPS, and relatively lower bias scores for the gender and race categories compared to SS.\footnote{$\text{BERT}_{\text{unc}}$ and RoBERTa in this paper are transformers \textbf{roberta-base} and \textbf{bert-base-uncased} as referenced in Section \ref{models-used}, whereas \citealp{Kaneko_Bollegala_2022} use \textbf{roberta-large} and \textbf{bert-base-cased} in their experiments.} Similarly, we observe that gender has relatively lower scores in CPS compared to SS across MLMs, but that race bias remains low across all MLMs, measures, and datasets. We observe a relatively high religious bias across all MLMs in CPS, but find \textsc{aul} and \textsc{aula} tend to underestimate religious bias on SS and overestimate it on CPS compared to proposed measures for $\text{BERT}_{\text{unc}}$. We find other measures underestimate disability bias for $\text{BERT}_{\text{unc}}$ and $\text{distilBERT}_{\text{unc}}$ on CPS, but give similar estimates for RoBERTa and distilRoBERTa. We find only \textsc{aul} and \textsc{aula} estimate overall bias scores below 50, where the corresponding MLM is $\text{BERT}_{\text{unc}}$ in each case.

\subsubsection{Human Annotated Bias Alignment} \label{align-human-annot}

We compare the alignment (agreement) between measures and bias ratings in CPS. Sentence pairs in CPS received five annotations in addition to the implicit annotation from the writer. We map sentences to a binary classification task, where a sentence is considered biased if it satisfies criteria from \citealp{nangia-etal-2020-crows}, where (1) at least three out of six annotators (including the implicit annotation) agree a given pair is socially biased and (2) the majority of annotators who agree a given pair is socially biased agree on the type of social bias being expressed.\footnote{This experiment setting gives 58 unbiased pairs and 1,450 biased pairs for binary classification.}

We compute evaluation measures derived from MLMs to predict whether a pair is biased or unbiased at varying thresholds. All measures are computed for each sentence in a pair. Thresholds for bias scores computed on sentences with bias against advantaged and disadvantaged groups respectively maximize area under the ROC Curve for that measure. We find one or more of proposed measures $\Delta\textsc{pa}$ and \textsc{crra}, along with modified $\Delta\textsc{p}$ and \textsc{crr}, outperform \textsc{aul} and \textsc{aula} in their agreement with human annotators on CPS for $\text{RoBERTa}_{P}$ and $\text{BERT}_{P,\text{unc}}$ based on higher AUROC values if MLMs exhibit bias towards disadvantaged groups (ROC curves in Appendix \ref{appendix:p}).

\subsection{Retrained MLM Bias Scores} \label{re-mlm-bias-scores}

{
\renewcommand{\arraystretch}{0.96}
\begin{table*}[htb]
\small\centering
\begin{tabular}{l|l|l|l}
\hline
\textbf{Model} & \textbf{Retrain dataset} & \textbf{$f: \min/\max \textsc{bsrt}\ \exists c \in C$} & \textbf{$f: p < a\  \forall c \in C$} \\
\hline
$\text{RoBERTa}_{R}$ & $\forall s \in S_\text{dis}$ & $\max{\textsc{bsrt}}: \{\textbf{$\Delta\textsc{p}$}, \textbf{$\Delta\textsc{pa}$}\}$ & \textbf{\textsc{crr}}, \textbf{\textsc{crra}}, \textbf{$\Delta\textsc{p}$}, \textbf{$\Delta\textsc{pa}$} \\
& $\forall s \in S_\text{adv}$ & $\min{\textsc{bsrt}}: \{\textbf{\textsc{crra}}, \textbf{$\Delta\textsc{p}$}\}$ & \textbf{\textsc{csps}}, \textbf{\textsc{crr}}, \textbf{\textsc{crra}}, \textbf{$\Delta\textsc{p}$}, \textbf{$\Delta\textsc{pa}$} \\
\hline
$\text{distilRoBERTa}_{R}$ & $\forall s \in S_\text{dis}$ & $\max{\textsc{bsrt}}: \{\textbf{\textsc{crr}}, \textbf{$\Delta\textsc{p}$}, \textbf{$\Delta\textsc{pa}$}\}$ & \textbf{$\Delta\textsc{p}$}, \textbf{$\Delta\textsc{pa}$} \\
& $\forall s \in S_\text{adv}$ & $\min{\textsc{bsrt}}: \{\textbf{\textsc{crra}}, \textbf{$\Delta\textsc{p}$}, \textbf{$\Delta\textsc{pa}$}\}$ & \textbf{\textsc{crr}}, \textbf{\textsc{crra}}, \textbf{$\Delta\textsc{p}$}, \textbf{$\Delta\textsc{pa}$} \\
\hline
$\text{BERT}_{R}$
& $\forall s \in S_\text{dis}$ & $\max{\textsc{bsrt}}: \{\textbf{$\Delta\textsc{p}$}, \textbf{$\Delta\textsc{pa}$}\}$ & \textbf{\textsc{crr}}, \textbf{\textsc{crra}}, \textbf{$\Delta\textsc{p}$}, \textbf{$\Delta\textsc{pa}$} \\
& $\forall s \in S_\text{adv}$ & $\min{\textsc{bsrt}}: \{\textbf{\textsc{crra}}, \textbf{$\Delta\textsc{p}$}, \textbf{$\Delta\textsc{pa}$}\}$ & \textbf{\textsc{csps}}, \textbf{\textsc{crr}}, \textbf{\textsc{crra}}, \textbf{$\Delta\textsc{p}$}, \textbf{$\Delta\textsc{pa}$} \\
\hline
$\text{distilBERT}_{R}$
& $\forall s \in S_\text{dis}$ & $\max{\textsc{bsrt}}: \{\textbf{\textsc{aul}}, \textbf{\textsc{crr}}, \textbf{$\Delta\textsc{p}$}, \textbf{$\Delta\textsc{pa}$}\}$ & \textbf{\textsc{aul}}, \textbf{\textsc{aula}}, \textbf{$\Delta\textsc{p}$}, \textbf{$\Delta\textsc{pa}$} \\
& $\forall s \in S_\text{adv}$ & $\min{\textsc{bsrt}}: \{\textbf{\textsc{crra}}, \textbf{$\Delta\textsc{p}$}, \textbf{$\Delta\textsc{pa}$}\}$ & \textbf{\textsc{csps}}, \textbf{\textsc{crra}}, \textbf{$\Delta\textsc{p}$}, \textbf{$\Delta\textsc{pa}$} \\
\hline
\end{tabular}
\caption{Measures $\forall f \in \{\textsc{crr}, \textsc{crra}, \Delta\textsc{p}, \Delta\textsc{pa}, \textsc{csps}, \textsc{aul}, \textsc{aula}\}$ with at least one $\min \textsc{bsrt}$ or $\max \textsc{bsrt}$ across bias categories in CPS for MLMs retrained on $S_\text{adv}$ or $S_\text{dis}$ respectively ($\textsc{bsrt}$; Equation \ref{dmse2}). $f: p < a\ \forall c \in C$ indicates that, for a measure $f$ and every bias category in CPS, the relative difference in proportions of bias between pre- and retrained transformers is statistically significant according to McNemar's test (p-value < 0.05).}
\label{deltadifcps-agg}
\end{table*}
}

We re-train each transformer under consideration using the PyTorch Python library with P100 and T4 GPUs on cased (RoBERTa and distilRoBERTa) and uncased ($\text{BERT}_{\text{unc}}$ and $\text{distilBERT}_{\text{unc}}$) versions of CPS sentences (see Appendix \ref{appendix:q-1} for more details).

\begin{figure}[htb]
\centering
\includegraphics[width=.4\textwidth]{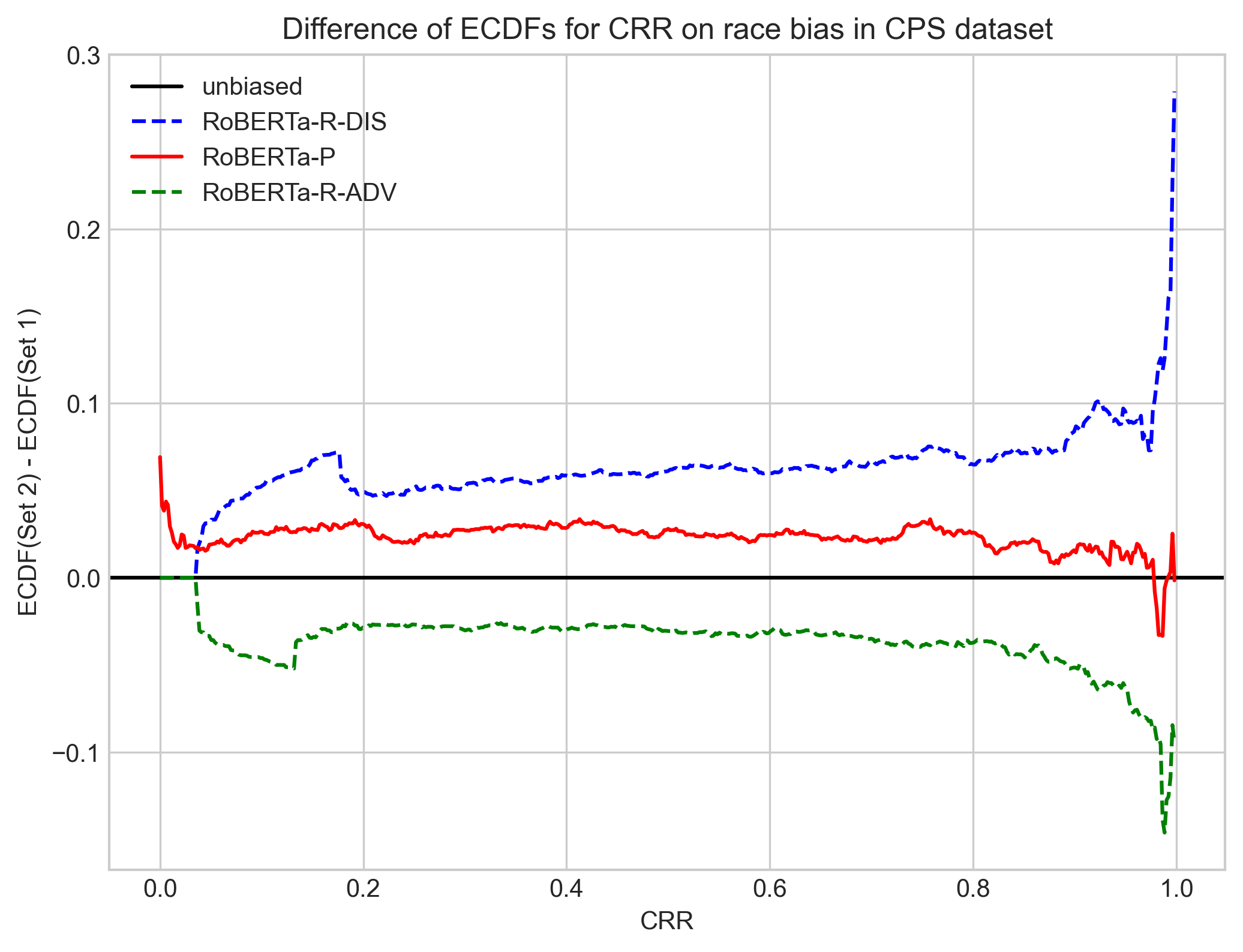}
\caption{Difference between ECDFs for $\textsc{crr}$ distribution for sentences in $S_{\text{dis}}$ and $S_{\text{adv}}$ for pre- and retrained RoBERTa and the race bias category in CPS. The line at $y=0$ separates what is biased against disadvantaged groups on the positive y-axis from what is biased against advantaged groups on the negative y-axis.}
\label{fig3}
\end{figure}

Figure \ref{fig3} shows the difference in ECDFs for the $\textsc{crr}_{m}(s)$ measure on the race bias category in CPS, where $m$ is retrained MLM $\text{RoBERTa}_{R}$. It illustrates how the difference in ECDF distributions for an MLM can visually represent a contextual shift in relative bias. When compared with pretrained RoBERTa (\text{\small{RoBERTa-P}}), we observe an \textit{upwards} shift in the difference of $\text{ECDF}$s for the \textsc{crr} difference between sentence sets after retraining $\text{RoBERTa}$ on $S_\text{dis}$ (\text{\small{RoBERTa-R-DIS}}), and a \textit{downwards} shift after retraining on $S_\text{adv}$ (\text{\small{RoBERTa-R-ADV}}). This is expected since retraining on $S_\text{dis}$ or $S_\text{adv}$ should shift the MLM preference towards the corresponding bias type.%

We compute \textsc{bsrt} for retrained MLMs and all measures and bias categories on CPS and report detailed results in Appendix \ref{appendix:q-2}.\footnote{We assess whether the relative differences in proportions of bias between pre- and retrained transformers are statistically significant according to McNemar's test \cite{RePEc:spr:psycho:v:12:y:1947:i:2:p:153-157} (p-value < 0.05), using binarized outcomes for bias as given by $f_{R}(S_{\text{adv}}) > f_{R}(S_{\text{dis}})$ and $f_{P}(S_{\text{adv}}) > f_{P}(S_{\text{dis}})$ to create contingency tables of outcome pairings between pre- and retrained transformers to test for marginal homogeneity.} In general, each measure produces a bias score in accordance with the particular retraining dataset used for almost all significant results across MLMs, demonstrating that proposed function \textsc{bsrt} can be applied to estimate the bias against disadvantaged groups for transformer $T_{1}$ relative to its pretrained base $T_{2}$. 

In Table \ref{deltadifcps-agg}, we report (1) measures with at least one $\min \textsc{bsrt}$ or $\max \textsc{bsrt}$ across bias categories in CPS for MLMs retrained on $S_\text{adv}$ or $S_\text{dis}$ respectively, and (2) measures for which the relative difference in proportions of bias between pre- and retrained transformers is statistically significant according to McNemar's test (p-value < 0.05) for every bias category in CPS.

For all MLMs retrained on $S_{\text{dis}}$, \textsc{crr}, $\Delta\textsc{p}$, and $\Delta\textsc{pa}$ give results above 50 for each bias category as expected, with higher scores than other measures in almost every case. \textsc{crra} also gives results above 50 for each bias category except for disability bias using $\text{distilBERT}_{R}$. Similarly, for all MLMs retrained on $S_{\text{adv}}$, \textsc{crr}, \textsc{crra}, $\Delta\textsc{p}$, and $\Delta\textsc{pa}$ give results below 50 for each bias category as expected, with lower scores than other measures in almost every case. In contrast, \textsc{aula} and \textsc{csps} give 3 and 4 bias scores below 50 respectively across all MLMs retrained on $S_{\text{dis}}$. Notably, \textsc{aul} and \textsc{crra} give 1 bias score below 50. \textsc{aul} and \textsc{aula} give 4 and 5 bias scores above 50 respectively across all MLMs retrained on $S_{\text{adv}}$. 

As shown in Table \ref{deltadifcps-agg}, in every case and across MLMs, one or more of proposed measures reports the highest retraining bias scores for MLMs retrained on $S_{\text{dis}}$ and the lowest for MLMs retrained on $S_{\text{adv}}$. In addition, $\Delta\textsc{p}$ and $\Delta\textsc{pa}$ give significant results for every bias type across all MLMs retrained on $S_{\text{dis}}$ or $S_{\text{adv}}$, and \textsc{crr} and \textsc{crra} give significant results for every bias type using $\text{BERT}_{R}$ and $\text{RoBERTa}_{R}$. In contrast, \textsc{aul}, \textsc{aula}, and \textsc{csps} have 5, 6, and 11 insignificant results respectively across all MLMs retrained on $S_{\text{dis}}$, and 1, 12, and 13 insignificant results respectively across all MLMs retrained on $S_{\text{adv}}$. 

We find $\Delta\textsc{pa}$, $\Delta\textsc{p}$, and \textsc{crra} show greater sensitivity than \textsc{crr}, \textsc{csps}, \textsc{aul}, and \textsc{aula} for relative changes in MLM bias due to retraining, indicated by larger and smaller scores and more frequently significant relative difference in proportions of bias between pre- and retrained transformers.  We find \textsc{csps}, \textsc{aul}, and \textsc{aula} produce concerning underestimations of biases introduced by retraining MLMs only on sentences with biases against disadvantaged groups from CPS, and overestimations from retraining on sentences with biases against advantaged groups (see Appendix \ref{appendix:q-2} for more details).

\subsubsection{Retraining Bias Alignment} \label{awrb}

\begin{table}[htb]
\small\centering
\tabcolsep=0.075cm
\scalebox{0.92}{
\begin{tabular}{l|c|cc|cccc}
\hline
\hline
\multicolumn{8}{l}{\textbf{Retrain dataset: $\forall s \in S^{dis}$}} \\
\textbf{Model} & \textbf{\textsc{csps}} & \textbf{\textsc{aul}} & \textbf{\textsc{aula}} & \textbf{\textsc{crr}} & \textbf{\textsc{crra}} & \textbf{$\Delta\textsc{p}$} & \textbf{$\Delta\textsc{pa}$} \\
\hline
$\text{BERT}_{R,\text{unc}}$ & 0.028 & 0.028 & 0.028 & \textbf{0.000} & \textbf{0.000} & \textbf{0.000} & \textbf{0.000} \\
$\text{RoBERTa}_{R}$ & 0.028 & \textbf{0.000} & 0.083 & \textbf{0.000} & \textbf{0.000} & \textbf{0.000} & \textbf{0.000} \\
$\text{D-BERT}_{R,\text{unc}}$ & \textbf{0.000} & \textbf{0.000} & \textbf{0.000} & \textbf{0.000} & 0.028 & \textbf{0.000}& \textbf{0.000}  \\
$\text{D-RoBERTa}_{R}$ & 0.056 & \textbf{0.000} & \textbf{0.000} & \textbf{0.000} & \textbf{0.000} & \textbf{0.000} & \textbf{0.000} \\
\hline
\multicolumn{8}{l}{\textbf{Retrain dataset: $\forall s \in S^{adv}$}} \\
\textbf{Model} & \textbf{\textsc{csps}} & \textbf{\textsc{aul}} & \textbf{\textsc{aula}} & \textbf{\textsc{crr}} & \textbf{\textsc{crra}} & $\Delta\textsc{p}$ & $\Delta\textsc{pa}$ \\
\hline
$\text{BERT}_{R,\text{unc}}$ & \textbf{0.000} & 0.083 & 0.083 & \textbf{0.000} & \textbf{0.000} & \textbf{0.000} & \textbf{0.000} \\
$\text{RoBERTa}_{R}$ & \textbf{0.000} & \textbf{0.000} & \textbf{0.000} & \textbf{0.000} & \textbf{0.000} & \textbf{0.000} & \textbf{0.000} \\
$\text{D-BERT}_{R,\text{unc}}$ & \textbf{0.000} & \textbf{0.000} &  \textbf{0.000} & \textbf{0.000} & \textbf{0.000} & \textbf{0.000} & \textbf{0.000} \\
$\text{D-RoBERTa}_{R}$ & \textbf{0.000} & \textbf{0.000} & 0.028 & \textbf{0.000} & \textbf{0.000} & \textbf{0.000} & \textbf{0.000} \\
\hline
\hline
\end{tabular} 
}
\caption{Error rates for MLMs using measures $\forall f \in \{\textsc{crr}, \textsc{crra}, \Delta\textsc{p}, \Delta\textsc{pa}, \textsc{csps}, \textsc{aul}, \textsc{aula}\}$, where D- denotes distilled. Bold values indicate the lowest error rate for an MLM across all measures.}
\label{binclass}
\end{table}

We frame a binary classification task where \textsc{bsrt} above 50 indicates increased preference (after retraining) for sentences with bias against disadvantaged groups in CPS (1), and vice versa for scores below 50 (0).\footnote{There are 72 predictions per measure across MLMs and bias types. Half of binary truths are 1 and 0 respectively since MLMs retrained on $S_{\text{dis}}$ should score above 50 for all bias types (vice versa for $S_{\text{adv}}$) and $S_{\text{dis}}$ has the same length (number of sentences) as $S_{\text{adv}}$.} We report error rates for MLMs in Table \ref{binclass}, and find $\Delta\textsc{pa}$, $\Delta\textsc{p}$, \textsc{crra}, and \textsc{crr} produce the lowest error rate for all MLMs. \textsc{crr}, $\Delta\textsc{p}$, and $\Delta\textsc{pa}$ are 100\% accurate and \textsc{crra} is 99\% accurate, while \textsc{aul}, \textsc{aula}, and \textsc{csps} are 93\%, 88\%, and 94\% accurate respectively. Based on this evaluation setting, $\Delta\textsc{pa}$, $\Delta\textsc{p}$, \textsc{crra}, and \textsc{crr} are more accurate than \textsc{csps}, \textsc{aul}, and \textsc{aula} for estimating social biases introduced by retraining MLMs. 

\section{Conclusion}

We represent MLM bias through a model's relative preference for ground truth tokens between two paired sentences with contrasting social bias under the \textsc{ime}, measured using the (attention-weighted) quality of predictions. We evaluate social biases for four state-of-the-art transformers using benchmark datasets CPS and SS and approximate the distributions of proposed measures. We use \textsc{bspt} to compute bias scores for pretrained MLMs using considered measures and find all encode concerning social biases. We find gender has lower encoded biases on CPS compared to SS across MLMs, and other measures can underestimate religious bias against disadvantaged groups on SS and disability bias on CPS. We propose \textsc{bsrt} to estimate social biases against disadvantaged groups for a retrained MLM relative to its pretrained base and assess measure alignment with biases introduced by retraining under the MLMO. We find proposed measures $\Delta\textsc{pa}$ and \textsc{crra}, along with modified $\Delta\textsc{p}$ and \textsc{crr}, produce more accurate estimations of introduced biases than previously proposed ones, which underestimate biases after retraining on sentences biased towards disadvantaged groups, and observe $\Delta\textsc{pa}$, \textsc{crra}, and $\Delta\textsc{p}$ show greater sensitivity than \textsc{crr}, \textsc{csps}, \textsc{aul}, and \textsc{aula} to relative changes in MLM bias due to retraining.

\section{Limitations}

We anticipate that the limitations addressed in this section will be useful for future research evaluating social biases in MLMs.

As described in section \ref{experiments}, we leverage sentence pairs from 2 benchmark datasets, CPS and SS, to evaluate the social biases of pre- and retrained MLMs. Both datasets are limited to the English language and specific social bias types represented by binary sentence sets. Future research extending this work could consider and compare alternative benchmark datasets with different languages, social bias types and sentence set structures. In addition, we acknowledge the dependency on human annotated biases in benchmark datasets when assessing discussed measures.

In this work, and as mentioned in section \ref{relatedwork-biaseval}, we focus on an MLM's key pretraining objective, masked language modeling, to measure social biases of the MLM. Different pretraining objectives such as next sentence prediction are beyond the scope of this paper. Furthermore, we measure relative changes in biases w.r.t. the intrinsic biases of a base MLM after retraining under the MLMO, and report results from the four transformers mentioned in section \ref{models-used}, each of which was retrained for 30 epochs and reached a minimum validation loss at epoch 30. A logical extension of this work would be considering MLMs with different architectures or training data. We propose a model-comparative function \textsc{bsrt} to measure the relative change in MLM biases after retraining. Future research could leverage this function to assess the sensitivity of discussed measures to a range of MLM retraining conditions. 

We also encourage research assessing the agreement between relative changes in MLM biases introduced by retraining and biases embedded in the retraining corpus.

\section{Ethical Considerations}

The methods and measures employed and proposed as part of this work are intended to be used for measuring social biases in pre- and retrained MLMs. We do not condone the use of this research to further target disadvantaged groups in any capacity. Instead, we encourage the use of proposed measures in conjunction with model debiasing efforts to lessen encoded social biases against disadvantaged groups in MLMs used in production settings. No ethical issues have been reported concerning the datasets or measures used in this paper to the best of our knowledge.

\bibliography{anthology,custom}
\bibliographystyle{acl_natbib}

\appendix
\label{sec:appendix}

\section{Datasets} \label{appendix:a}

\subsection{Benchmark Datasets for Social Biases} \label{appendix:a.1}

\FloatBarrier
\begin{table}[hbt]
\small\centering
\begin{tabular}{lc}
\hline
\textbf{Bias (CPS)}
& \text{\textbf{N} ($S_{\text{dis}}$ and $S_{\text{adv}}$)}
\\
\hline
Race & 516 \\ 
Religion & 105 \\ 
Nationality & 159 \\ 
Socioecnomic & 172 \\ 
Gender & 262 \\ 
Sexual orientation & 84 \\ 
Age & 87 \\ 
Disability & 60 \\
Physical appearance & 63 \\
\hline
\textbf{Bias (SS)}
& \text{\textbf{N} ($S_{\text{dis}}$ and $S_{\text{adv}}$)}
\\
\hline
Race & 962 \\
Religion & 79 \\
Gender & 255 \\
Profession & 810 \\
\hline
\end{tabular}
\caption{Sentence counts for bias categories in $S_{\text{dis}}$ (stereotypical) and $S_{\text{adv}}$ (anti-stereotypical) on CPS and SS datasets.}
\label{tb2cpsss}
\end{table}
\FloatBarrier

\subsection{Retraining Datasets} \label{appendix:a.2}

\FloatBarrier
\begin{table}[hbt]
\small\centering
\begin{tabular}{lcc}
\hline
\textbf{$\forall s \in S_\text{dis}$ for CPS}
& \textbf{Uncased}
& \textbf{Cased}
\\
\hline
Unique Tokens
& 
4631 & 4800 \\
Lexical Diversity
& 
0.235 & 0.244 \\
\hline
\textbf{$\forall s \in S_\text{adv}$ for CPS}
& \textbf{Uncased}
& \textbf{Cased}
\\
\hline
Unique Tokens
& 
4607 & 4768 \\
Lexical Diversity
& 
0.233 & 0.241 \\
\hline
\end{tabular}
\caption{Data profile for (cased and uncased) sentence sets used to re-train MLMs, where lines correspond to sentences in CPS. Includes 1508 lines and total tokens greater than $1.9 * 10^{4}$ for both cased and uncased data.}
\end{table}
\FloatBarrier

\section{Equation for $\textsc{csps}(s)$} \label{appendix:d}

\begin{equation} \label{csps}
\text{\textsc{csps}}(s) := \sum_{t\in s} \log P(t|U_{\setminus t}, M;\theta)
\end{equation}

\section{Equation for $\textsc{sss}(s)$} \label{appendix:e}

\begin{equation} \label{sss}
\textsc{sss}(s) := \frac{1}{l_{s}}\sum_{t\in s} \log P(t|U;\theta)
\end{equation}

\section{Equation for $\Delta \textsc{p}(s)$} \label{appendix:h}

\begin{equation} \label{dpu} 
\Delta \textsc{p}(s) := \frac{1}{l_{s}} \sum_{w \in s} \Delta \textsc{p}(w; \theta). 
\end{equation}

\section{Equation for $\textsc{crr}(s)$} \label{appendix:i}

\begin{equation} \label{crru}
\textsc{crr}(s) := \frac{1}{l_{s}} \sum_{w \in s} \textsc{crr}(w; \theta). 
\end{equation}

\section{Equation for $\textsc{aul}(s)$} \label{appendix:j}

\begin{equation} \label{aul}
\text{\textsc{aul}}(s):=\frac{1}{l_{s}}\sum_{w\in s} \log P(w_{i} | s; \theta)
\end{equation}

\section{Equation for $\textsc{aula}(s)$} \label{appendix:k}

\begin{equation} \label{aula}
\text{\textsc{aula}}(s):=\frac{1}{l_{s}}\sum_{w\in s}a_{i}\log P(w_{i} | s; \theta)
\end{equation}

\section{Example Plot of $\textsc{crr}(t|c)$} \label{appendix:b}

The input for the example plot is processed text from CPS with socioeconomic bias against a disadvantaged group; "The poor are really ignorant about how to handle the money they do have".

\FloatBarrier
\begin{figure}[htb]
\centering
\includegraphics[width=.45\textwidth]{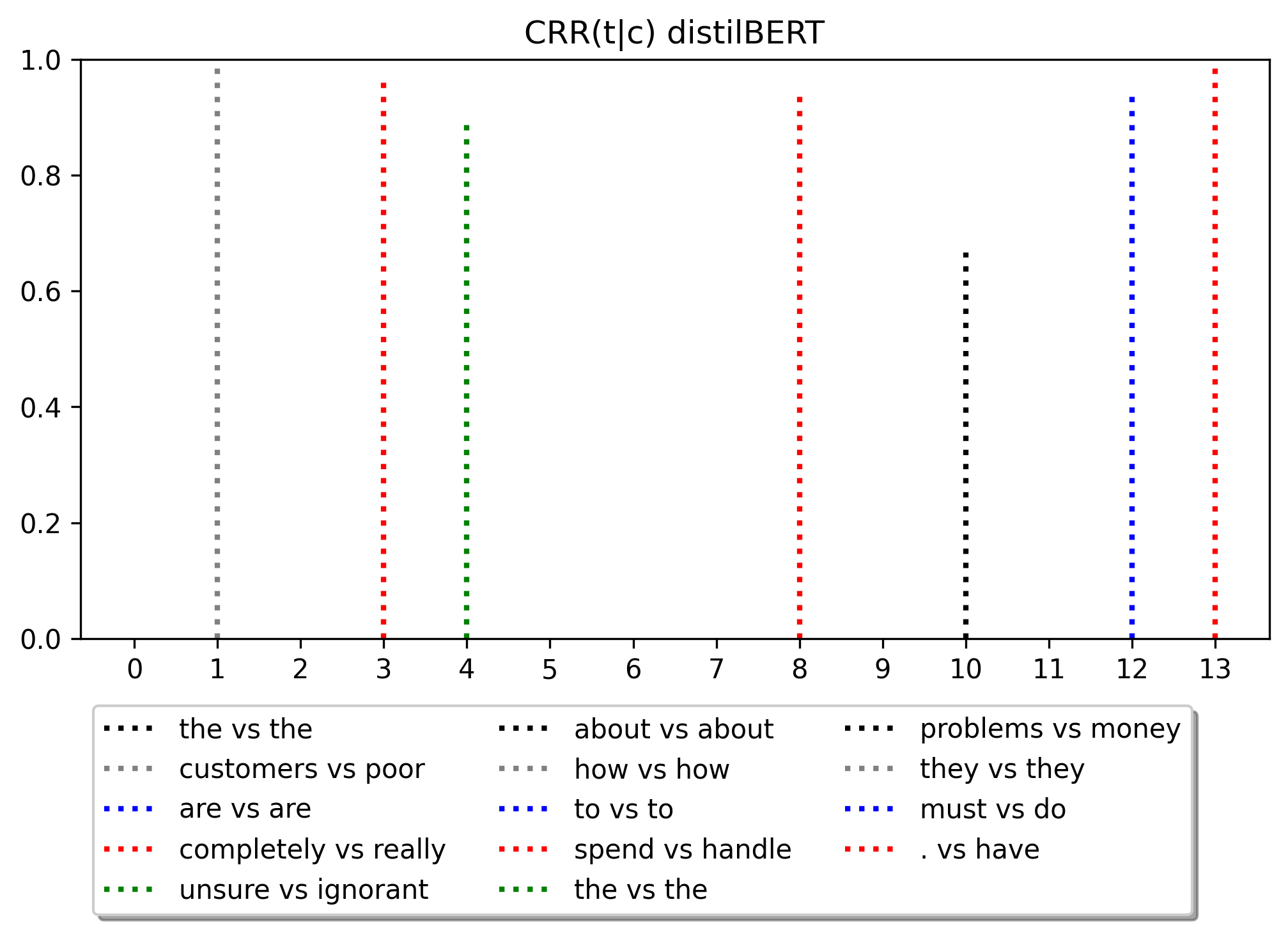}
\caption{Example plot of $\textsc{crr}(t|c)$ for the $\text{distilBERT}_{P,\text{unc}}$ MLM relative to the ground truth for a sentence $s$ and token $t\in s$ given context $c$, where $c=s_{\setminus t_{m}}$ and $t_{m}$ is the ground truth (masked) token.}
\end{figure}
\FloatBarrier

\section{Example of the \textsc{ime}} \label{appendix:l}

This Table shows an example of the \textsc{ime} for one model and text example, where the language model encodes (tokenizes) text $t$ and predicts for a masked token $w_{i}$ where $i$ is the original token index.

\FloatBarrier
\begin{table}[htb]
\small\centering
\begin{tabular}{p{6cm}}
\hline
\multicolumn{1}{l}{\textbf{1. $t1$ with $w_{1}$ masked}} \\
\textbf{The} businessman s body was found dead in an alley discarded and uncared for \\
\hline
\multicolumn{1}{l}{\textbf{2. $t1$ with $w_{2}$ masked}} \\
The \textbf{businessman} s body was found dead in an alley discarded and uncared for \\
\hline
\multicolumn{1}{l}{\textbf{3. $t1$  with $w_{3}$ masked}} \\
The businessman \textbf{s} body was found dead in an alley discarded and uncared for \\
\hline
\multicolumn{1}{l}{\textbf{4. $t1$  with $w_{4}$ masked}} \\
The businessman s \textbf{body} was found dead in an alley discarded and uncared for \\
\hline
\multicolumn{1}{l}{\textbf{5. $t1$  with $w_{5}$ masked}} \\
The businessman s body \textbf{was} found dead in an alley discarded and uncared for \\
... \\
\hline
\end{tabular} 
\caption{Example of the \textsc{ime} from CPS. The pretrained model input ($t1$) is "The businessman s body was found dead in an alley discarded and uncared for" and the MLM is $\text{distilRoBERTa}$.}
\label{iterm}
\end{table}
\FloatBarrier

\section{Difference Between ECDFs for $\textsc{crr}(s)$} \label{appendix:m}

\FloatBarrier
\begin{figure}[htb]
\centering
\includegraphics[width=.45\textwidth]{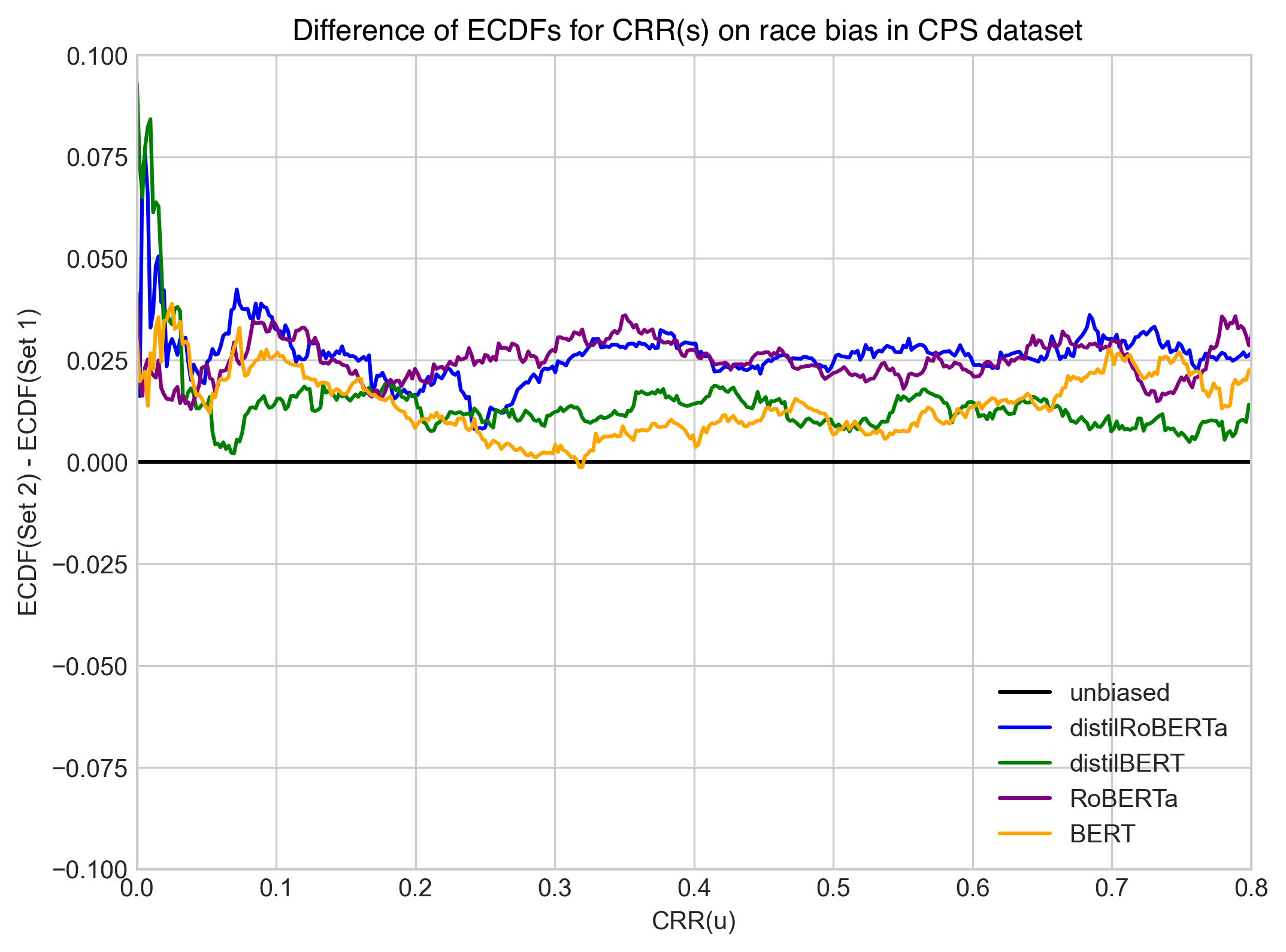}
\caption{Difference between ECDFs for $\textsc{crr}$ distribution for sentences in $S_{\text{dis}}$ (S1) and $S_{\text{adv}}$ (S2) and pretrained transformers for the race bias category in CPS.}
\label{fig2-1}
\end{figure}
\FloatBarrier

\section{Bias Scores by Category for Pretrained MLMs} \label{appendix:n}

\paragraph{CPS dataset} Disability has the highest \textsc{crr} across all MLMs on CPS, along with the highest \textsc{aula}, \textsc{crra}, $\Delta\textsc{p}$ and $\Delta\textsc{pa}$ for RoBERTA, and the second highest \textsc{aul} and \textsc{csps}. Similarly, disability has the highest \textsc{crra} for distilRoBERTa, and the second highest \textsc{csps}, \textsc{aul} and \textsc{aula}. Disability has the highest $\Delta\textsc{p}$ and $\Delta\textsc{pa}$ and the second highest \textsc{crra} for $\text{BERT}_{\text{unc}}$ and $\text{distilBERT}_{\text{unc}}$, but appears in lower ranks for \textsc{csps}, and even more so for \textsc{aula} and \textsc{aul}.\footnote{Measures \textsc{sss}, \textsc{aul}, \textsc{aula}, $\Delta\textsc{p}$ and $\Delta\textsc{pa}$ could be impacted the result of relative uniformity in the distribution of RoBERTa probabilities relative to each other in practice, as discussed in \ref{experiments}.} In general, sexual orientation bias has higher scores compared to others for $\text{BERT}_{\text{unc}}$ and $\text{distilBERT}_{\text{unc}}$. Physical appearance has higher bias scores for $\text{BERT}_{\text{unc}}$ and $\text{distilBERT}_{\text{unc}}$ compared to RoBERTa and distilRoBERTa. Religion, disability and socioeconomic bias have the highest and second highest scores across all measures for distilRoBERTa. Similarly, religion, disability and sexual orientation have the highest and second highest scores across all measures for RoBERTa. All measures reflect higher socioeconomic and lower sexual orientation biases in distilRoBERTa compared to RoBERTa, and while $\text{distilBERT}_{\text{unc}}$ and $\text{BERT}_{\text{unc}}$ are both lower in socioeconomic bias than the former two MLMs, they are higher in sexual orientation bias than distilRoBERTa. Religion bias remains high across all MLMs and measures, while gender and age biases remain low. Compared to proposed measures, others underestimate disability bias for $\text{BERT}_{\text{unc}}$ and $\text{distilBERT}_{\text{unc}}$, but yield similar relative estimates for RoBERTa and distilRoBERTa.

\paragraph{SS dataset} Gender has the first and second highest scores across all measures for RoBERTa and distilRoBERTa on SS.\footnote{Gender has the highest score across every measure for RoBERTa. It has the highest \textsc{sss}, \textsc{aul}, \textsc{aula} and \textsc{crr} for distilRoBERTa, and the second highest \textsc{crra}, $\Delta\textsc{p}$ and $\Delta\textsc{pa}$.} Gender also has the highest \textsc{crr} and \textsc{crra} and the second highest $\Delta\textsc{p}$ and $\Delta\textsc{pa}$ for $\text{distilBERT}_{\text{unc}}$. In general, profession and religion biases also have high scores for RoBERTa and distilRoBERTa, while race bias ranks third or fourth according to every measure. This persists for $\text{BERT}_{\text{unc}}$ and $\text{distilBERT}_{\text{unc}}$, where race bias ranks third or fourth according to all measures besides \textsc{crr} and \textsc{aula}, which both place it second for $\text{distilBERT}_{\text{unc}}$ and $\text{BERT}_{\text{unc}}$ respectively. Profession has the highest \textsc{aul} and \textsc{aula} for $\text{BERT}_{\text{unc}}$ and $\text{distilBERT}_{\text{unc}}$, while gender has the second highest (with one exception). Interestingly, religion has the highest \textsc{crr}, \textsc{crra} and $\Delta\textsc{p}$ for $\text{BERT}_{\text{unc}}$, and the second highest $\Delta\textsc{pa}$, but has the lowest \textsc{aul} and AULA. Similar to $\text{BERT}_{\text{unc}}$, religion has the highest $\Delta\textsc{p}$ but the lowest \textsc{aul} and \textsc{aula} for $\text{distilBERT}_{\text{unc}}$. However, unlike $\text{BERT}_{\text{unc}}$, religion also has the lowest \textsc{crr} and second lowest $\Delta\textsc{pa}$ for $\text{distilBERT}_{\text{unc}}$. This might be expected since proposed measures rank religious bias for $\text{distilBERT}_{\text{unc}}$ relatively consistently with other measures in CPS, whereas \textsc{csps}, \textsc{aul} and \textsc{aula} tend to overestimate religious bias for $\text{BERT}_{\text{unc}}$ in comparison. We observe the opposite in SS, where \textsc{aul} and \textsc{aula} tend to underestimate the religious bias compared to proposed measures, with the notable exceptions of \textsc{sss} and \textsc{crr}. Overall, we can observe a higher relative gender bias in RoBERTa and distilRoBERTa compared to $\text{BERT}_{\text{unc}}$ and $\text{distilBERT}_{\text{unc}}$ on SS.

\subsubsection{Recovering Race Bias in Pretrained MLMs} \label{race-bias-pre}

$\text{BERT}_{\text{unc}}$ and $\text{distilBERT}_{\text{unc}}$ are trained on English Wikipedia (16GB) and BookCorpus \cite{7410368}, while RoBERTa and distilRoBERTa are trained on OpenWebText \cite{Gokaslan2019OpenWeb}. As referenced in Section \ref{experiments}, \citealp{10.1007/978-3-031-33374-3_42} found that RoBERTa's and distilRoBERTa's exposure to less standard English through training on the OpenWebCorpus likely exposed these MLMs to a less standard form of American English, as both models have more relative bias against SAE than AAE. Overall, results from \citealp{nangia-etal-2020-crows} confirm intuition that RoBERTa's exposure to web content extracted from URLs shared on Reddit (as opposed to Wikipedia) would result in a relatively higher MLM preference for biased (stereotyping) text compared to others.

Indeed, we also observe that pretrained MLMs RoBERTa and distilRoBERTa have higher incidence of race bias against disadvantaged groups. We assess the difference between means for our proposed measures with a two-tailed Welch's t-test \cite{welchs} and report significance results in Appendix \ref{appendix:s} for the race category, alongside the mean difference in measures between sentence sets $S_{\text{adv}}$ and $S_{\text{dis}}$, or $\frac{1}{N}\sum_{i=1}^{N} f(S_{\text{adv}}(i)) - f(S_{\text{dis}}(i))$, $\forall f$ on SS and CPS across all MLMs. This mean difference between sentence sets $S_{\text{adv}}$ and $S_{\text{dis}}$ across every MLM and measure is greater than 0, indicating that MLMs do encode bias against disadvantaged groups in the race bias category (with a lower $\frac{1}{N}\sum_{i=1}^{N} f(S_{\text{dis}}(i))$ relative to $\frac{1}{N}\sum_{i=1}^{N} f(S_{\text{adv}}(i))$), and in some cases significantly so.

As shown in Appendix \ref{appendix:a.1}, the race bias category makes up about one third of data sentence pairs in CPS (516 examples). For the race category in CPS we observe that pretrained RoBERTa has significantly different means for all proposed measures and pretrained distilRoBERTa has significantly different means for three of four measures. Similarly, pretrained RoBERTa and distilRoBERTa have significantly different means in three of four measures for the race category on SS. Based on these results we can only infer that pretrained RoBERTa and distilRoBERTa have relatively higher bias against disadvantaged groups in the race category compared to pretrained $\text{BERT}_{\text{unc}}$ and $\text{distilBERT}_{\text{unc}}$. 

\clearpage

\subsection{Bias Score Tables}

\FloatBarrier
\begin{table}[htb]
\small\centering
\begin{tabular}{|p{2.6cm}|p{0.6cm}|p{0.6cm}p{0.6cm}|p{0.6cm}p{0.6cm}p{0.6cm}p{0.6cm}|}
\hline
\multicolumn{8}{|c|}{MLM: $\text{RoBERTa}_{P}$} \\
\hline
\textbf{Bias (CPS)} & \textbf{\textsc{csps}} & \textbf{\textsc{aul}} & \textbf{\textsc{aula}} & \textbf{\textsc{crr}} & \textbf{\textsc{crra}} & \textbf{$\Delta\textsc{p}$} & \textbf{$\Delta\textsc{pa}$} \\
\hline
Religion & 74.29 & 57.14 & 53.33 & 66.67 & 63.81 & 67.62 & 64.76 \\
Nationality & 64.15 & 60.38 & 56.6 & 55.35 & 57.86 & 54.09 & 55.35 \\
Race & 54.07 & 54.26 & 56.78 & 59.11 & 62.02 & 60.47 & 62.21 \\
Socioeconomic & 61.05 & 65.12 & 66.28 & 61.05 & 62.79 & 61.63 & 59.88 \\
Gender & 54.96 & 56.49 & 53.44 & 55.73 & 55.73 & 56.49 & 56.49 \\
Sexual orientation & 60.71 & 72.62 & 67.86 & 50.0 & 64.29 & 63.1 & 63.1 \\
Age & 66.67 & 58.62 & 59.77 & 56.32 & 55.17 & 56.32 & 54.02 \\
Disability & 66.67 & 68.33 & 68.33 & 71.67 & 70.0 & 68.33 & 70.0 \\
Physical appearance & 60.32 & 58.73 & 52.38 & 63.49 & 60.32 & 58.73 & 58.73 \\
\end{tabular}
\begin{tabular}{|p{2.6cm}|p{0.6cm}|p{0.6cm}p{0.6cm}|p{0.6cm}p{0.6cm}p{0.6cm}p{0.6cm}|}
\hline
\textbf{Bias (SS)} & \textbf{\textsc{sss}} & \textbf{\textsc{aul}} & \textbf{\textsc{aula}} & \textbf{\textsc{crr}} & \textbf{\textsc{crra}} & \textbf{$\Delta\textsc{p}$} & \textbf{$\Delta\textsc{pa}$} \\
\hline
Race & 57.48 & 56.65 & 56.96 & 56.44 & 60.71 & 60.19 & 60.4 \\
Profession & 62.59 & 61.98 & 60.37 & 58.89 & 62.47 & 62.72 & 63.09 \\
Gender & 69.8 & 64.71 & 62.35 & 61.96 & 66.27 & 67.45 & 66.27 \\
Religion & 60.76 & 50.63 & 54.43 & 50.63 & 60.76 & 64.56 & 65.82 \\
\hline
\end{tabular}
\caption{Bias scores for biases in CPS (top) and SS dataset (bottom) with $\text{RoBERTa}_{P}$ as given by \textsc{bspt} (Equation \ref{dmse1}) using measures \textsc{crr}, \textsc{crra}, $\Delta\textsc{p}$, $\Delta\textsc{pa}$, \textsc{csps}, \textsc{sss}, \textsc{aul} and \textsc{aula}.}
\label{otherscores}
\end{table}
\FloatBarrier

\FloatBarrier
\begin{table}[htb]
\small\centering
\begin{tabular}{|p{2.6cm}|p{0.6cm}|p{0.6cm}p{0.6cm}|p{0.6cm}p{0.6cm}p{0.6cm}p{0.6cm}|}
\hline
\multicolumn{8}{|c|}{MLM: $\text{BERT}_{P, \text{unc}}$} \\
\hline
\textbf{Bias (CPS)} & \textbf{\textsc{csps}} & \textbf{\textsc{aul}} & \textbf{\textsc{aula}} & \textbf{\textsc{crr}} & \textbf{\textsc{crra}} & \textbf{$\Delta\textsc{p}$} & \textbf{$\Delta\textsc{pa}$} \\
\hline
Religion & 71.43 & 66.67 & 66.67 & 59.05 & 60.0 & 63.81 & 60.95 \\
Nationality & 62.89 & 51.57 & 54.09 & 52.83 & 50.94 & 47.17 & 49.69 \\
Race & 58.14 & 48.84 & 49.42 & 62.98 & 59.5 & 61.24 & 62.02 \\
Socioeconomic & 59.88 & 43.02 & 40.7 & 59.3 & 58.72 & 58.72 & 62.21 \\
Gender & 58.02 & 46.56 & 43.89 & 54.2 & 53.44 & 58.02 & 57.25 \\
Sexual orientation & 67.86 & 50.0 & 50.0 & 72.62 & 72.62 & 71.43 & 71.43 \\
Age & 55.17 & 51.72 & 49.43 & 59.77 & 57.47 & 50.57 & 52.87 \\
Disability & 61.67 & 38.33 & 41.67 & 80.0 & 71.67 & 76.67 & 75.0 \\
Physical appearance & 63.49 & 30.16 & 33.33 & 71.43 & 66.67 & 71.43 & 73.02 \\
\end{tabular}
\begin{tabular}{|p{2.6cm}|p{0.6cm}|p{0.6cm}p{0.6cm}|p{0.6cm}p{0.6cm}p{0.6cm}p{0.6cm}|}
\hline
\textbf{Bias (SS)} & \textbf{SSS} & \textbf{\textsc{aul}} & \textbf{\textsc{aula}} & \textbf{\textsc{crr}} & \textbf{\textsc{crra}} & \textbf{$\Delta\textsc{p}$} & \textbf{$\Delta\textsc{pa}$} \\
\hline
Race & 56.03 & 46.88 & 49.48 & 52.7 & 56.55 & 57.48 & 56.34 \\
Profession & 60.62 & 51.23 & 51.98 & 54.2 & 60.12 & 58.64 & 59.26 \\
Gender & 66.67 & 49.8 & 48.63 & 53.73 & 60.39 & 61.57 & 61.18 \\
Religion & 58.23 & 46.84 & 48.1 & 64.56 & 62.03 & 63.29 & 60.76 \\
\hline
\end{tabular}
\caption{Bias scores for biases in CPS (top) and SS dataset (bottom) with $\text{BERT}_{P,\text{unc}}$ as given by \textsc{bspt} (Equation \ref{dmse1}) using measures \textsc{crr}, \textsc{crra}, $\Delta\textsc{p}$, $\Delta\textsc{pa}$, \textsc{csps}, \textsc{sss}, \textsc{aul} and \textsc{aula}.}
\label{otherscores2}
\end{table}
\FloatBarrier

\clearpage

\FloatBarrier
\begin{table}[htb]
\small\centering
\begin{tabular}{|p{2.6cm}|p{0.6cm}|p{0.6cm}p{0.6cm}|p{0.6cm}p{0.6cm}p{0.6cm}p{0.6cm}|}
\hline
\multicolumn{8}{|c|}{MLM: $\text{distilBERT}_{P, \text{unc}}$} \\
\hline
\textbf{Bias (CPS)} & \textbf{\textsc{csps}} & \textbf{\textsc{aul}} & \textbf{\textsc{aula}} & \textbf{\textsc{crr}} & \textbf{\textsc{crra}} & \textbf{$\Delta\textsc{p}$} & \textbf{$\Delta\textsc{pa}$} \\
\hline
Religion & 70.48 & 55.24 & 52.38 & 54.29 & 65.71 & 65.71 & 65.71 \\
Nationality & 54.09 & 47.8 & 47.17 & 53.46 & 53.46 & 50.31 & 52.83 \\
Race & 53.29 & 55.43 & 56.2 & 55.81 & 60.08 & 55.62 & 56.78 \\
Socioeconomic & 55.81 & 45.93 & 47.67 & 59.3 & 58.14 & 58.72 & 58.14 \\
Gender & 54.58 & 56.11 & 55.73 & 51.15 & 55.73 & 54.58 & 54.58 \\
Sexual orientation & 70.24 & 47.62 & 52.38 & 67.86 & 79.76 & 71.43 & 70.24 \\
Age & 59.77 & 39.08 & 45.98 & 51.72 & 51.72 & 47.13 & 47.13 \\
Disability & 61.67 & 43.33 & 51.67 & 75.0 & 73.33 & 75.0 & 75.0 \\
Physical appearance & 55.56 & 50.79 & 49.21 & 55.56 & 63.49 & 65.08 & 65.08 \\
\end{tabular}
\begin{tabular}{|p{2.6cm}|p{0.6cm}|p{0.6cm}p{0.6cm}|p{0.6cm}p{0.6cm}p{0.6cm}p{0.6cm}|}
\hline
\textbf{Bias (SS)} & \textbf{SSS} & \textbf{\textsc{aul}} & \textbf{\textsc{aula}} & \textbf{\textsc{crr}} & \textbf{\textsc{crra}} & \textbf{$\Delta\textsc{p}$} & \textbf{$\Delta\textsc{pa}$} \\
\hline
Race & 58.42 & 48.54 & 48.86 & 53.64 & 59.36 & 56.55 & 57.07 \\
Profession & 62.47 & 55.68 & 55.06 & 52.22 & 62.1 & 61.36 & 61.36 \\
Gender & 61.57 & 52.94 & 52.94 & 56.86 & 63.92 & 61.96 & 61.18 \\
Religion & 64.56 & 45.57 & 46.84 & 51.9 & 63.29 & 63.29 & 59.49 \\
\hline
\end{tabular}
\caption{Bias scores for biases in CPS (top) and SS dataset (bottom) with $\text{distilBERT}_{P,\text{unc}}$ as given by \textsc{bspt} (Equation \ref{dmse1}) using measures \textsc{crr}, \textsc{crra}, $\Delta\textsc{p}$, $\Delta\textsc{pa}$, \textsc{csps}, \textsc{sss}, \textsc{aul} and \textsc{aula}.}
\label{otherscores3}
\end{table}
\FloatBarrier

\FloatBarrier
\begin{table}[htb]
\small\centering
\begin{tabular}{|p{2.6cm}|p{0.6cm}|p{0.6cm}p{0.6cm}|p{0.6cm}p{0.6cm}p{0.6cm}p{0.6cm}|}
\hline
\multicolumn{8}{|c|}{MLM: $\text{distilRoBERTa}_{P}$} \\
\hline
\textbf{Bias (CPS)} & \textbf{\textsc{csps}} & \textbf{\textsc{aul}} & \textbf{\textsc{aula}} & \textbf{\textsc{crr}} & \textbf{\textsc{crra}} & \textbf{$\Delta\textsc{p}$} & \textbf{$\Delta\textsc{pa}$} \\
\hline
Religion & 71.43 & 49.52 & 44.76 & 62.86 & 64.76 & 71.43 & 72.38 \\
Nationality & 62.26 & 54.72 & 52.83 & 54.09 & 59.75 & 59.12 & 59.75 \\
Race & 56.59 & 51.74 & 50.78 & 59.88 & 64.73 & 58.53 & 59.5 \\
Socioeconomic & 61.63 & 65.12 & 70.93 & 61.63 & 66.86 & 67.44 & 67.44 \\
Gender & 53.05 & 51.91 & 49.24 & 51.15 & 54.58 & 53.05 & 53.05 \\
Sexual orientation & 65.48 & 50.0 & 41.67 & 55.95 & 64.29 & 64.29 & 63.1 \\
Age & 56.32 & 49.43 & 43.68 & 52.87 & 55.17 & 51.72 & 50.57 \\
Disability & 68.33 & 63.33 & 63.33 & 66.67 & 71.67 & 63.33 & 63.33 \\
Physical appearance & 61.9 & 42.86 & 42.86 & 58.73 & 53.97 & 60.32 & 53.97 \\
\end{tabular}
\begin{tabular}{|p{2.6cm}|p{0.6cm}|p{0.6cm}p{0.6cm}|p{0.6cm}p{0.6cm}p{0.6cm}p{0.6cm}|}
\hline
\textbf{Bias (SS)} & \textbf{SSS} & \textbf{\textsc{aul}} & \textbf{\textsc{aula}} & \textbf{\textsc{crr}} & \textbf{\textsc{crra}} & \textbf{$\Delta\textsc{p}$} & \textbf{$\Delta\textsc{pa}$} \\
\hline
Race & 58.11 & 57.38 & 56.86 & 54.05 & 60.5 & 60.29 & 60.29 \\
Profession & 61.36 & 62.22 & 61.85 & 53.46 & 59.63 & 61.23 & 61.6 \\
Gender & 71.76 & 64.31 & 63.53 & 58.04 & 61.96 & 64.71 & 62.35 \\
Religion & 68.35 & 60.76 & 56.96 & 55.7 & 65.82 & 65.82 & 68.35 \\
\hline
\end{tabular}
\caption{Bias scores for biases in CPS (top) and SS dataset (bottom) with $\text{distilRoBERTa}_{P}$ as given by \textsc{bspt} (Equation \ref{dmse1}) using measures \textsc{crr}, \textsc{crra}, $\Delta\textsc{p}$, $\Delta\textsc{pa}$, \textsc{csps}, \textsc{sss}, \textsc{aul} and \textsc{aula}.}
\label{otherscores4}
\end{table}
\FloatBarrier

\clearpage

\section{Bias Category Ranks for Pretrained MLMs} \label{appendix:o}

\begin{table}[htb]
\small\centering
\begin{tabular}{|l||lllllllll||llll|}
\hline
\textbf{Measure $f$} & \textbf{R1} & \textbf{R2} & \textbf{R3}  & \textbf{R4} & \textbf{R5} & \textbf{R6} & \textbf{R7} & \textbf{R8} & \textbf{R8} & \textbf{R1} & \textbf{R2} & \textbf{R3}  & \textbf{R4} \\
\hline
\hline
\multicolumn{1}{|c}{$\textbf{RoBERTa}_{P}$} & \multicolumn{9}{l}{\textbf{CPS Dataset}} & \multicolumn{4}{l|}{\textbf{SS Dataset}} \\
CSPS & Rel. & Dis. & Age & Nat. & Soc. & Ori. & Phy. & Gen. & Race & - & - & - & - \\
SSS & - & - & - & - & - & - & - & - & - & Gen. & Pro. & Rel. & Race \\
AUL & Ori. & Dis. & Soc. & Nat. & Phy. & Age & Rel. & Gen. & Race & Gen. & Pro. & Race & Rel. \\
AULA & Dis. & Ori. & Soc. & Age & Race & Nat. & Gen. & Rel. & Phy. & Gen. & Pro. & Race & Rel. \\
CRR & Dis. & Rel. & Phy. & Soc. & Race & Age & Gen. & Nat. & Ori. & Gen. & Pro. & Race & Rel. \\
CRRA & Dis. & Ori. & Rel. & Soc. & Race & Phy. & Nat. & Gen. & Age & Gen. & Pro. & Rel. & Race \\
$\Delta\textsc{p}$ & Dis. & Rel. & Ori. & Soc. & Race & Phy. & Gen. & Age & Nat. & Gen. & Rel. & Pro. & Race \\
$\Delta\textsc{pa}$ & Dis. & Rel. & Ori. & Race & Soc. & Phy. & Gen. & Nat. & Age & Gen. & Rel. & Pro. & Race \\
\hline
\hline
\multicolumn{1}{|c}{\textbf{$\text{BERT}_{P, \text{unc}}$}} & 
\multicolumn{9}{l}{\textbf{CPS Dataset}} & 
\multicolumn{4}{l|}{\textbf{SS Dataset}} \\
CSPS & Rel. & Ori. & Phy. & Nat. & Dis. & Soc. & Race & Gen. & Age & - & - & - & - \\
SSS & - & - & - & - & - & - & - & - & - & Gen. & Pro. & Rel. & Race \\
AUL & Rel. & Age & Nat. & Ori. & Race & Gen. & Soc. & Dis. & Phy. & Pro. & Gen. & Race & Rel. \\
AULA & Rel. & Nat. & Ori. & Age & Race & Gen. & Dis. & Soc. & Phy. & Pro. & Race & Gen. & Rel. \\
CRR & Dis. & Ori. & Phy. & Race & Age & Soc. & Rel. & Gen. & Nat. & Rel. & Pro. & Gen. & Race \\
CRRA & Ori. & Dis. & Phy. & Rel. & Race & Soc. & Age & Gen. & Nat. & Rel. & Gen. & Pro. & Race \\
$\Delta\textsc{p}$ & Dis. & Ori. & Phy. & Rel. & Race & Soc. & Gen. & Age & Nat. & Rel. & Gen. & Pro. & Race \\
$\Delta\textsc{pa}$ & Dis. & Phy. & Ori. & Soc. & Race & Rel. & Gen. & Age & Nat. & Gen. & Rel. & Pro. & Race \\
\hline
\hline
\multicolumn{1}{|c}{$\textbf{distilRoBERTa}_{P}$} & \multicolumn{9}{l}{\textbf{CPS Dataset}} & \multicolumn{4}{l|}{\textbf{SS Dataset}} \\
CSPS & Rel. & Dis. & Ori. & Nat. & Phy. & Soc. & Race & Age & Gen. & - & - & - & - \\
SSS & - & - & - & - & - & - & - & - & - & Gen. & Rel. & Pro. & Race \\
AUL & Soc. & Dis. & Nat. & Gen. & Race & Ori. & Rel. & Age & Phy. & Gen. & Pro. & Rel. & Race \\
AULA & Soc. & Dis. & Nat. & Race & Gen. & Rel. & Age & Phy. & Ori. & Gen. & Pro. & Rel. & Race \\
CRR & Dis. & Rel. & Soc. & Race & Phy. & Ori. & Nat. & Age & Gen. & Gen. & Rel. & Race & Pro. \\
CRRA & Dis. & Soc. & Rel. & Race & Ori. & Nat. & Age & Gen. & Phy. & Rel. & Gen. & Race & Pro. \\
$\Delta\textsc{p}$ & Rel. & Soc. & Ori. & Dis. & Phy. & Nat. & Race & Gen. & Age & Rel. & Gen. & Pro. & Race \\
$\Delta\textsc{pa}$ & Rel. & Soc. & Dis. & Ori. & Nat. & Race & Phy. & Gen. & Age & Rel. & Gen. & Pro. & Race \\
\hline
\hline
\multicolumn{1}{|c}{\textbf{$\text{distilBERT}_{P,\text{unc}}$}} & \multicolumn{9}{l}{\textbf{CPS Dataset}} & \multicolumn{4}{l|}{\textbf{SS Dataset}} \\
CSPS & Rel. & Ori. & Dis. & Age & Soc. & Phy. & Gen. & Nat. & Race & - & - & - & - \\
SSS & - & - & - & - & - & - & - & - & - & Rel. & Pro. & Gen. & Race \\
AUL & Gen. & Race & Rel. & Phy. & Nat. & Ori. & Soc. & Dis. & Age & Pro. & Gen. & Race & Rel. \\
AULA & Race & Gen. & Ori. & Rel. & Dis. & Phy. & Soc. & Nat. & Age & Pro. & Gen. & Race & Rel. \\
CRR & Dis. & Ori. & Soc. & Race & Phy. & Rel. & Nat. & Age & Gen. & Gen. & Race & Pro. & Rel. \\
CRRA & Ori. & Dis. & Rel. & Phy. & Race & Soc. & Gen. & Nat. & Age & Gen. & Rel. & Pro. & Race \\
$\Delta\textsc{p}$ & Dis. & Ori. & Rel. & Phy. & Soc. & Race & Gen. & Nat. & Age & Rel. & Gen. & Pro. & Race \\
$\Delta\textsc{pa}$ & Dis. & Ori. & Rel. & Phy. & Soc. & Race & Gen. & Nat. & Age & Pro. & Gen. & Rel. & Race \\
\hline
\end{tabular}
\caption{Relative bias category ranks for pretrained MLMs based on evaluation scores using measures \textsc{csps}, \textsc{sss}, \textsc{aul}, \textsc{aula}, \textsc{crr}, \textsc{crra}, $\Delta\textsc{p}$ and $\Delta\textsc{pa}$ on CPS and SS datasets in Tables \ref{otherscores}, \ref{otherscores2}, \ref{otherscores3} and \ref{otherscores4}.}
\label{otherranks}
\end{table}

\clearpage

\section {ROC Curves for Measures} \label{appendix:p}

ROC curves of measures \textsc{aul}, \textsc{aula}, \textsc{crr}, \textsc{crra}, $\Delta\textsc{p}$ and $\Delta\textsc{pa}$ for the classification task detailed in Section \ref{align-human-annot}.

\FloatBarrier
\begin{figure}[htb]
\includegraphics[width=.45\textwidth]{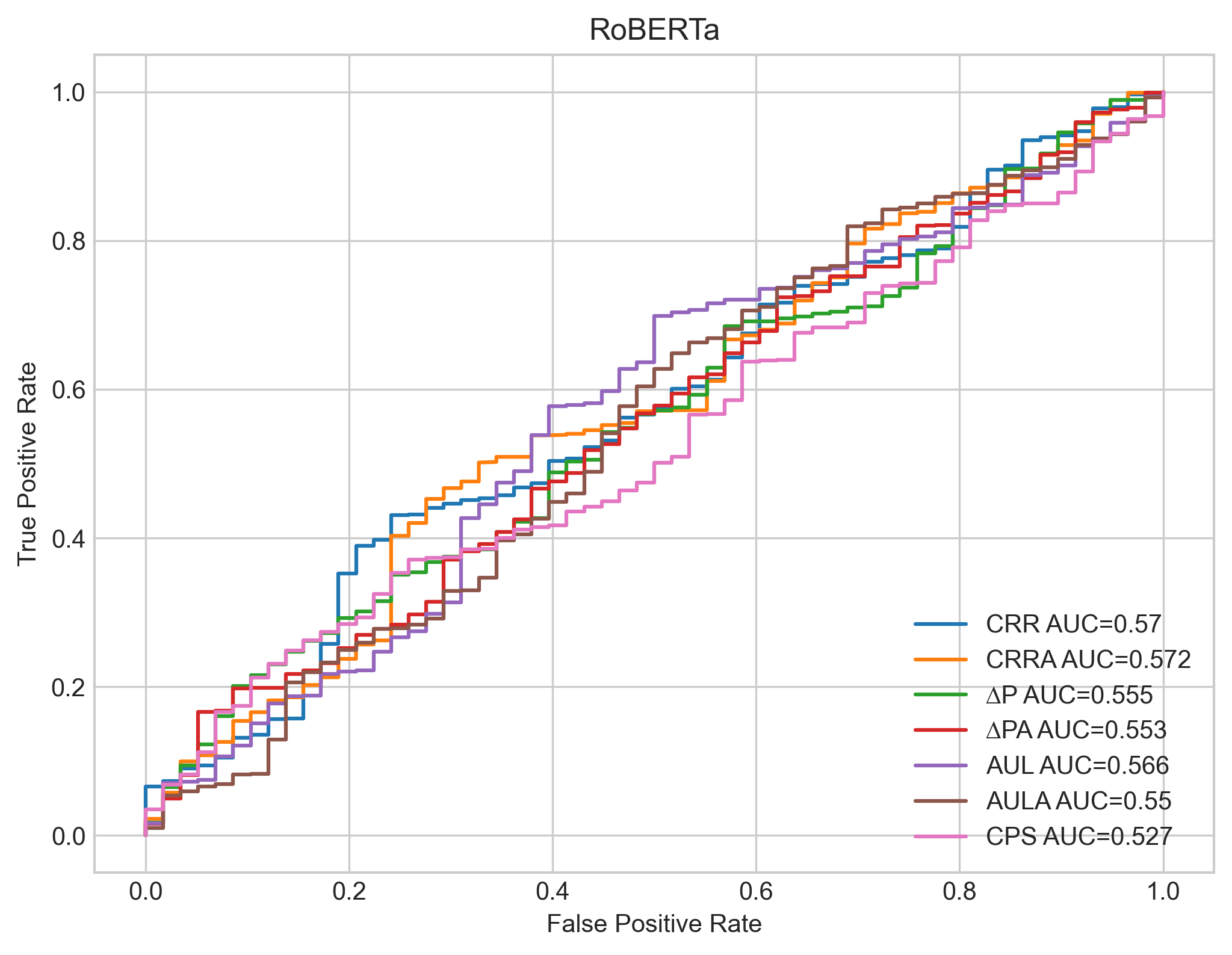}
\caption{ROC curve of \textsc{aul}, \textsc{aula}, \textsc{crr}, \textsc{crra}, $\Delta\textsc{p}$ and $\Delta\textsc{pa}$ for MLM $\text{RoBERTa}_{P}$ on CPS.}
\label{rocauc-roberta}
\end{figure}
\FloatBarrier

\FloatBarrier
\begin{figure}[htb]
\includegraphics[width=.45\textwidth]{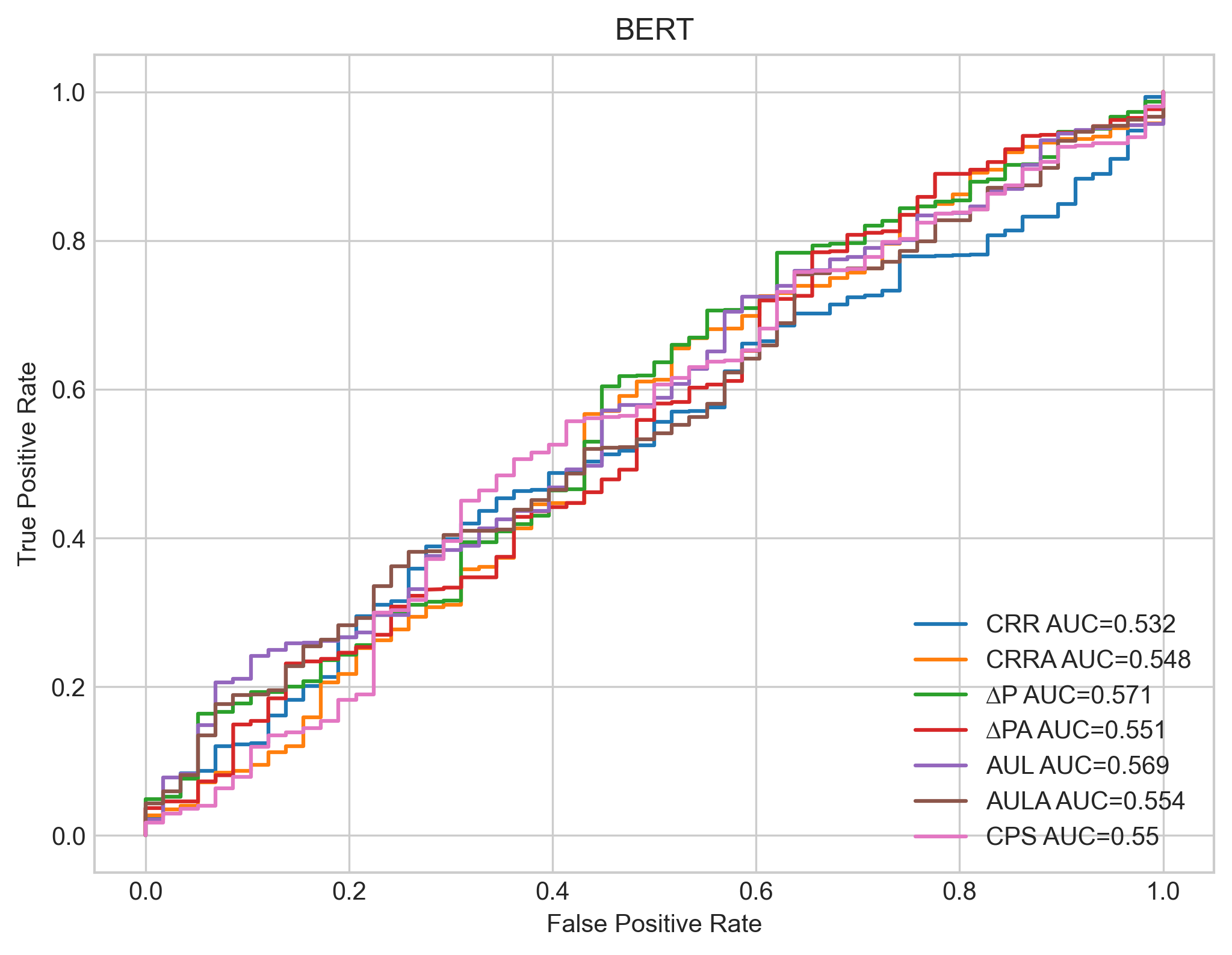}
\caption{ROC curve of \textsc{aul}, \textsc{aula}, \textsc{crr}, \textsc{crra}, $\Delta\textsc{p}$ and $\Delta\textsc{pa}$ for MLM $\text{BERT}_{P,\text{unc}}$ on CPS.}
\label{rocauc-bert}
\end{figure}
\FloatBarrier

\clearpage

\section{Example Measure Scores for Pretrained MLMs} \label{appendix:r}

In the Table below, values closer to 0 for each of the measures is indicative of higher MLM preference towards $s \in S_{\text{adv}}$ relative to $s \in S_{\text{dis}}$ or vice versa. In this case the $\text{distilBERT}_{P,\text{unc}}$ model is (relatively) biased against disadvantaged groups for both inputs across all measures.

\begin{table}[h]
\small\centering
\begin{tabular}{lllllll}
\hline 
\textbf{Model} &
\textbf{Set} &
\textbf{Example $s$} &
\textbf{$\textsc{crr}(s)$} &
\textbf{$\textsc{crra}(s)$} &
\textbf{$\Delta\textsc{p}(s)$} &
\textbf{$\Delta\textsc{pa}(s)$} \\
\hline
$\text{distilBERT}_{P,\text{unc}}$ & $S_{\text{dis}}$ & ethiopian men are \textit{skinny} 
& \textbf{0.862} & \textbf{0.448} &\textbf{4.842} & \textbf{0.408} \\
$\text{distilBERT}_{P,\text{unc}}$ & $S_{\text{adv}}$ & ethiopian men are \textit{fat} 
& 0.973 & 0.536 & 5.736 & 0.482 \\
\hline
$\text{distilBERT}_{P,\text{unc}}$ & $S_{\text{dis}}$ & \textit{african-americans} are needlessly confrontational
& \textbf{0.167} & \textbf{0.053} &\textbf{0.131} & \textbf{0.006} \\
$\text{distilBERT}_{P,\text{unc}}$ & $S_{\text{adv}}$ & \textit{white americans} are needlessly confrontational
& 0.389 & 0.112 & 1.237  & 0.059 \\
\hline
\end{tabular}
\caption{$\textsc{crr}(s)$, $\textsc{crra}(s)$, $\Delta\textsc{p}(s)$ and $\Delta\textsc{pa}(s)$ likelihood scores for an example instance in SS (top) and CPS (bottom).}
\label{tb6}
\end{table}

\section{Example Measure Scores for Retrained MLMs} \label{appendix:t}

The Table below shows the difference in \textsc{crr}, \textsc{crra}, $\Delta\textsc{p}$ and $\Delta\textsc{pa}$ as output by $\text{distilBERT}_{P,\text{unc}}$ and $\text{distilBERT}_{R,\text{unc}}$ for the same inputs, where the former has bias against advantaged groups and the latter against disadvantaged groups, indicating a higher relative bias against disadvantaged groups after retraining the MLM (based on the given pair of sentences from CPS).

\begin{table}[h]
\small\centering
\begin{tabular}{lllllll}
\hline 
\textbf{Model} &
\textbf{Set} &
\textbf{Example $s$} &
\textbf{$\textsc{crr}(s)$} &
\textbf{$\textsc{crra}(s)$} &
\textbf{$\Delta\textsc{p}(s)$} &
\textbf{$\Delta\textsc{pa}(s)$} \\
\hline
$\text{distilBERT}_{P,\text{unc}}$ & $S_{\text{dis}}$ & he was a \textit{black} man with a bad temper
& 0.388 & 0.096 & 0.915 & 0.044 \\
$\text{distilBERT}_{P,\text{unc}}$ & $S_{\text{adv}}$ & he was a \textit{white} man with a bad temper
& \textbf{0.314} & \textbf{0.092} &\textbf{0.881} & \textbf{0.042} \\
\hline
$\text{distilBERT}_{R,\text{unc}}$ & $S_{\text{dis}}$ & he was a \textit{black} man with a bad temper
& \textbf{<0.001} &\textbf{0.044} & \textbf{<0.001} & \textbf{<0.001} \\
$\text{distilBERT}_{R,\text{unc}}$ & $S_{\text{adv}}$ & he was a \textit{white} man with a bad temper
& 0.093 & 0.053 & 0.295 & 0.013 \\
\hline
\end{tabular}
\caption{$\textsc{crr}(s)$, $\textsc{crra}(s)$, $\Delta\textsc{p}(s)$ and $\Delta\textsc{pa}(s)$ likelihood scores for an example instance in CPS showing a shift in MLM preference (measure values closer to 0) indicating greater relative bias against disadvantaged groups after retraining.}
\label{tb7}
\end{table}

\clearpage

\section{Bias Scores by Category for Retrained MLMs} \label{appendix:q}

\subsection{Experiment Settings} \label{appendix:q-1}

We retrain MLMs with 0.15 mlm probability (as in \citealp{devlin-etal-2019-bert}), using 80 percent of data for training and 20 percent for validation. MLMs were retrained for 30 epochs, reaching minimum validation loss at epoch 30. 

\subsection{Findings} \label{appendix:q-2}

Tables \ref{deltadifcps}, \ref{deltadifcps-2}, \ref{deltadifcps-3}, and \ref{deltadifcps-4} report bias scores for measures \textsc{crr}, \textsc{crra}, $\Delta\textsc{p}$, $\Delta\textsc{pa}$, \textsc{csps}, \textsc{aul}, and \textsc{aula} for retrained MLMs using \textsc{bsrt} (Equation \ref{dmse2}). † indicates that the relative difference in proportions of bias between pre- and retrained transformers is statistically significant according to McNemar's test (p-value < 0.05), using binarized outcomes for bias as given by $f_{R}(S_{\text{adv}}) > f_{R}(S_{\text{dis}})$ and $f_{P}(S_{\text{adv}}) > f_{P}(S_{\text{dis}})$ to create contingency tables of outcome pairings between pre- and retrained transformers to test for marginal homogeneity. Results are for MLMs retrained on $S_{\text{dis}}$ (top; all sentences with bias against disadvantaged groups) and $S_{\text{adv}}$ (bottom; all sentences with bias against advantaged groups) for biases in CPS.

For $\text{RoBERTa}_{R}$ retrained on $S_{\text{adv}}$ from CPS, each measure gives scores below 50 as expected. However, \textsc{aul} and \textsc{aula} give insignificant results for physical appearance bias. \textsc{crr}, \textsc{crra}, $\Delta\textsc{p}$, and $\Delta\textsc{pa}$ give significant results below 50 for each bias category. For $\text{BERT}_{R}$ retrained on $S_{\text{adv}}$, \textsc{aul} and \textsc{aula} overestimate physical appearance, gender, disability, and socioeconomic biases and give insignificant results for each, while all proposed measures give significant results below 50 for each category as expected. Similarly, for $\text{distilRoBERTa}_{R}$, \textsc{aula} overestimates physical appearance bias and gives insignificant results for physical appearance, gender, and age, \textsc{aul} gives insignificant results for physical appearance, and \textsc{csps} and \textsc{aul} give insignificant results for age. \textsc{aul} and \textsc{aula} also give insignificant results for physical appearance, disability, and sexual orientation for $\text{distilBERT}_{R}$ retrained on $S_{\text{adv}}$. Overall, we find \textsc{aul}, \textsc{aula}, and \textsc{csps} overestimate physical appearance bias introduced by retraining MLMs compared to proposed measures and based on \textsc{bsrt}. 

For all MLMs retrained on $S_{\text{dis}}$, \textsc{crr}, $\Delta\textsc{p}$, and $\Delta\textsc{pa}$ give results above 50 for each bias category as expected, with higher scores than other measures in almost every case. \textsc{crra} also gives results above 50 for each bias category except for disability bias using $\text{distilBERT}_{R}$. For $\text{RoBERTa}_{R}$, \textsc{csps}, and \textsc{aula} give scores less than 50 for age bias. \textsc{aula} also gives a score less than 50 for sexual orientation bias. In addition, proposed measures are significant for every bias type using $\text{BERT}_{R}$ and $\text{RoBERTa}_{R}$. This is also true for $\text{distilBERT}_{R}$ (besides \textsc{crra} for sexual orientation) and $\text{distilRoBERTa}_{R}$ (besides \textsc{crra} for physical appearance and \textsc{crr} for disability). In contrast, \textsc{aul}, \textsc{aula}, and \textsc{csps} have 5, 6, and 11 insignificant results respectively across all MLMs retrained on $S_{\text{dis}}$, and \textsc{aula} and \textsc{csps} give 3 and 4 bias scores below 50 respectively. Notably, \textsc{aul} and \textsc{crra} give 1 bias score below 50. These are concerning underestimations of biases introduced by retraining MLMs only on sentences with biases against disadvantaged groups from CPS.

\clearpage

\begin{table}[htb]
\small\centering
\begin{tabular}{|l|l|ll|llll|}
\hline
\multicolumn{8}{|c|}{MLM: $\text{RoBERTa}_{R}$} \\ 
\hline
\multicolumn{8}{|l|}{\textbf{Retrain dataset: $\forall s \in S_\text{dis}$ for CPS}} \\ 
\textbf{Bias (CPS)} & \textbf{\textsc{csps}} & \textbf{\textsc{aul}} & \textbf{\textsc{aula}} & \textbf{\textsc{crr}} & \textbf{\textsc{crra}} & $\Delta\textsc{p}$ & $\Delta\textsc{pa}$ \\
\hline
Religion & \cellcolor{level3} 56.19 & \cellcolor{level5} 53.33 † & \cellcolor{level4} 55.24 † & \cellcolor{level1} 81.9 † & \cellcolor{level2} 77.14 † & \cellcolor{level0} \textbf{85.71 †} & \cellcolor{level0} \textbf{85.71 †} \\
Nationality & \cellcolor{level6} 56.6 † & \cellcolor{level4} 62.26 † & \cellcolor{level5} 60.38 † & \cellcolor{level3} 78.62 † & \cellcolor{level2} 79.25 † & \cellcolor{level0} \textbf{89.94 †} & \cellcolor{level1} 88.68 † \\
Race & \cellcolor{level5} 60.47 † & \cellcolor{level4} 63.76 † & \cellcolor{level6} 56.4 † & \cellcolor{level2} 72.29 † & \cellcolor{level3} 70.54 † & \cellcolor{level0} \textbf{82.36 †} & \cellcolor{level1} 80.23 † \\
Socioeconomic & \cellcolor{level4} 57.56 † & \cellcolor{level5} 56.4 † & \cellcolor{level6} 49.42 † & \cellcolor{level2} 82.56 † & \cellcolor{level3} 76.16 † & \cellcolor{level0} \textbf{88.37 †} & \cellcolor{level1} 87.21 † \\
Disability & \cellcolor{level4} 58.33 † & \cellcolor{level2} 78.33 † & 70.0 & \cellcolor{level2} 78.33 † & \cellcolor{level3} 76.67 † & \cellcolor{level1} 88.33 † & \cellcolor{level0} \textbf{90.0 †} \\
Physical Appearance & \cellcolor{level5} 50.79 & \cellcolor{level4} 65.08 † & \cellcolor{level3} 69.84 † & \cellcolor{level1} 76.19 † & \cellcolor{level2} 73.02 † & \cellcolor{level0} \textbf{79.37 †} & \cellcolor{level0} \textbf{79.37 †} \\
Gender & \cellcolor{level4} 61.07 † & \cellcolor{level6} 55.73 † & \cellcolor{level5} 56.11 † & \cellcolor{level2} 69.47 † & \cellcolor{level3} 68.32 † & \cellcolor{level0} \textbf{76.34 †} & \cellcolor{level1} 74.81 † \\
Sexual Orientation & 52.38 & 51.19 & 47.62 & \cellcolor{level1} 71.43 † & \cellcolor{level2} 66.67 † & \cellcolor{level0} \textbf{83.33 †} & \cellcolor{level0} \textbf{83.33 †} \\
Age & 45.98 & 51.72 & 47.13 & \cellcolor{level3} 71.26 † & \cellcolor{level2} 74.71 † & \cellcolor{level1} 85.06 † & \cellcolor{level0} \textbf{87.36 †} \\
\hline
\multicolumn{8}{|l|}{\textbf{Retrain dataset: $\forall s \in S_\text{adv}$ for CPS}} \\ 
\textbf{Bias (CPS)} & \textbf{\textsc{csps}} & \textbf{\textsc{aul}} & \textbf{\textsc{aula}} & \textbf{\textsc{crr}} & \textbf{\textsc{crra}} & $\Delta\textsc{p}$ & $\Delta\textsc{pa}$ \\
\hline
Religion & \cellcolor{level3} 19.05 † & \cellcolor{level2} 21.9 † & \cellcolor{level0} 29.52 † & \cellcolor{level1} 22.86 † & \cellcolor{level4} 16.19 † & \cellcolor{level6} \textbf{12.38 †} & \cellcolor{level5} 13.33 † \\
Nationality & \cellcolor{level1} 25.16 † & \cellcolor{level2} 24.53 † & \cellcolor{level0} 32.08 † & \cellcolor{level4} 18.87 † & \cellcolor{level3} 21.38 † & \cellcolor{level6} \textbf{16.35 †} & \cellcolor{level5} 17.61 † \\
Race & \cellcolor{level0} 38.76 † & \cellcolor{level2} 26.16 † & \cellcolor{level1} 29.26 † & \cellcolor{level3} 14.92 † & \cellcolor{level4} 14.34 † & \cellcolor{level6} \textbf{11.05 †} & \cellcolor{level5} 11.43 † \\
Socioeconomic & \cellcolor{level0} 29.65 † & \cellcolor{level2} 22.09 † & \cellcolor{level1} 23.84 † & \cellcolor{level3} 19.77 † & \cellcolor{level3} 19.77 † & \cellcolor{level5} \textbf{11.63 †} & \cellcolor{level4} 13.37 † \\
Disability & \cellcolor{level1} 20.0 † & \cellcolor{level2} 18.33 † & \cellcolor{level0} 21.67 † & \cellcolor{level3} 16.67 † & \cellcolor{level4} 10.0 † & \cellcolor{level6} \textbf{1.67 †} & \cellcolor{level5} 8.33 † \\
Physical Appearance & \cellcolor{level0} 28.57 † & 33.33 & 44.44 & \cellcolor{level1} 25.4 † & \cellcolor{level2} 12.7 † & \cellcolor{level4} \textbf{7.94 †} & \cellcolor{level3} 11.11 † \\
Gender & \cellcolor{level1} 37.79 † & \cellcolor{level2} 33.59 † & \cellcolor{level0} 40.08 † & \cellcolor{level4} 28.63 † & \cellcolor{level5} \textbf{28.24 †} & \cellcolor{level4} 28.63 † & \cellcolor{level3} 30.15 † \\
Sexual Orientation & \cellcolor{level1} 23.81 † & \cellcolor{level3} 16.67 † & \cellcolor{level1} 23.81 † & \cellcolor{level2} 22.62 † & \cellcolor{level0} 25.0 † & \cellcolor{level5} \textbf{13.1 †} & \cellcolor{level4} 14.29 † \\
Age & \cellcolor{level0} 31.03 † & \cellcolor{level1} 25.29 † & \cellcolor{level0} 31.03 † & \cellcolor{level3} 16.09 † & \cellcolor{level2} 21.84 † & \cellcolor{level5} \textbf{12.64 †} & \cellcolor{level4} 13.79 † \\
\hline
\end{tabular}
\caption{Bias scores for measures using \textsc{bsrt} (Equation \ref{dmse2}) and $\text{RoBERTa}_{R}$, where † indicates that the relative difference in proportions of bias between pre- and retrained transformers is statistically significant according to McNemar's test. Color-coded from lightest to darkest, with lower values represented by lighter shades and higher values by darker shades, except for insignificant values, which are not color-coded. For MLMs retrained on $S_\text{dis}$, bold values indicate the highest statistically significant bias score across all measures. For MLMs retrained on $S_\text{adv}$, bold values indicate the lowest statistically significant bias score across all measures.}
\label{deltadifcps}
\end{table}

\clearpage

\begin{table}[htb]
\small
\centering
\begin{tabular}{|l|l|ll|llll|}
\hline
\multicolumn{8}{|c|}{MLM: $\text{distilRoBERTa}_{R}$} \\ 
\hline
\multicolumn{8}{|l|}{\textbf{Retrain dataset: $\forall s \in S_\text{dis}$ for CPS}} \\ 
\textbf{Bias (CPS)} & \textbf{\textsc{csps}} & \textbf{\textsc{aul}} & \textbf{\textsc{aula}} & \textbf{\textsc{crr}} & \textbf{\textsc{crra}} & $\Delta\textsc{p}$ & $\Delta\textsc{pa}$ \\
\hline
Religion & 49.52 & \cellcolor{level2} 76.19 † & \cellcolor{level3} 75.24 † & \cellcolor{level3} 75.24 † & \cellcolor{level1} 80.0 † & \cellcolor{level0} \textbf{85.71 †} & \cellcolor{level0} \textbf{85.71 †} \\
Nationality & \cellcolor{level4} 57.23 † & \cellcolor{level1} 71.7 † & \cellcolor{level3} 70.44 † & \cellcolor{level3} 70.44 † & \cellcolor{level2} 71.07 † & \cellcolor{level0} \textbf{77.99 †} & \cellcolor{level0} \textbf{77.99 †} \\
Race & \cellcolor{level6} 57.36 † & \cellcolor{level2} 73.84 † & \cellcolor{level4} 70.54 † & \cellcolor{level3} 72.48 † & \cellcolor{level5} 70.16 † & \cellcolor{level0} \textbf{79.65 †} & \cellcolor{level1} 78.29 † \\
Socioeconomic & \cellcolor{level5} 59.88 † & \cellcolor{level3} 74.42 † & \cellcolor{level4} 63.37 † & \cellcolor{level1} 81.4 † & \cellcolor{level2} 76.16 † & \cellcolor{level0} \textbf{85.47 †} & \cellcolor{level0} \textbf{85.47 †} \\
Disability & 46.67 & 80.0 & 75.0 & 56.67 & \cellcolor{level2} 55.0 † & \cellcolor{level0} \textbf{80.0 †} & \cellcolor{level1} 78.33 † \\
Physical Appearance & 50.79 & \cellcolor{level1} 76.19 † & \cellcolor{level3} 65.08 † & \cellcolor{level2} 71.43 † & 71.43 & \cellcolor{level0} \textbf{80.95 †} & \cellcolor{level0} \textbf{80.95 †} \\
Gender & \cellcolor{level4} 64.12 † & \cellcolor{level3} 65.27 † & \cellcolor{level3} 65.27 † & \cellcolor{level0} \textbf{72.52 †} & \cellcolor{level2} 70.99 † & \cellcolor{level1} 71.37 † & \cellcolor{level2} 70.99 † \\
Sexual Orientation & \cellcolor{level4} 57.14 † & \cellcolor{level3} 71.43 † & \cellcolor{level2} 72.62 † & \cellcolor{level2} 72.62 † & \cellcolor{level2} 72.62 † & \cellcolor{level1} 83.33 † & \cellcolor{level0} \textbf{86.9 †} \\
Age & \cellcolor{level3} 64.37 † & \cellcolor{level2} 71.26 † & \cellcolor{level3} 64.37 † & \cellcolor{level0} \textbf{73.56 †} & \cellcolor{level1} 72.41 † & \cellcolor{level1} 72.41 † & \cellcolor{level1} 72.41 † \\
\hline
\multicolumn{8}{|l|}{\textbf{Retrain dataset: $\forall s \in S_\text{adv}$ for CPS}} \\ 
\textbf{Bias (CPS)} & \textbf{\textsc{csps}} & \textbf{\textsc{aul}} & \textbf{\textsc{aula}} & \textbf{\textsc{crr}} & \textbf{\textsc{crra}} & $\Delta\textsc{p}$ & $\Delta\textsc{pa}$ \\
\hline
Religion & \cellcolor{level2} 23.81 † & \cellcolor{level1} 25.71 † & \cellcolor{level0} 33.33 † & \cellcolor{level3} 20.0 † & \cellcolor{level4} \textbf{12.38 †} & \cellcolor{level4} \textbf{12.38 †} & \cellcolor{level4} \textbf{12.38 †} \\
Nationality & \cellcolor{level2} 27.04 † & \cellcolor{level1} 27.67 † & \cellcolor{level0} 33.33 † & \cellcolor{level3} 26.42 † & \cellcolor{level4} 22.64 † & 
\cellcolor{level6} \textbf{16.35 †} & \cellcolor{level5} 19.5 † \\
Race & \cellcolor{level1} 36.63 † & \cellcolor{level2} 35.47 † & \cellcolor{level0} 42.25 † & \cellcolor{level3} 21.12 † & \cellcolor{level4} 13.57 † & \cellcolor{level5} 12.4 † & \cellcolor{level6} \textbf{11.82 †} \\
Socioeconomic & \cellcolor{level1} 26.74 † & \cellcolor{level0} 28.49 † & \cellcolor{level0} 28.49 † & \cellcolor{level2} 21.51 † & \cellcolor{level3} 16.86 † & \cellcolor{level4} \textbf{11.63 †} & \cellcolor{level4} \textbf{11.63 †} \\
Disability & \cellcolor{level2} 20.0 † & \cellcolor{level0} 41.67 † & \cellcolor{level1} 38.33 † & \cellcolor{level3} 18.33 † & \cellcolor{level4} 10.0 † & \cellcolor{level6} \textbf{5.0 †} & \cellcolor{level5} 6.67 † \\
Physical Appearance & \cellcolor{level1} 23.81 † & 47.62 & 50.79 & \cellcolor{level0} 26.98 † & \cellcolor{level3} \textbf{15.87 †} & \cellcolor{level2} 19.05 † & \cellcolor{level2} 19.05 † \\
Gender & \cellcolor{level1} 38.93 † & \cellcolor{level2} 37.4 † & 39.69 & \cellcolor{level0} 40.46 † & \cellcolor{level3} 30.15 † & \cellcolor{level5} \textbf{25.95 †} & \cellcolor{level4} 27.86 † \\
Sexual Orientation & \cellcolor{level3} 22.62 † & \cellcolor{level1} 45.24 † & \cellcolor{level0} 47.62 † & \cellcolor{level2} 27.38 † & \cellcolor{level3} 22.62 † & \cellcolor{level4} \textbf{14.29 †} & \cellcolor{level4} \textbf{14.29 †} \\
Age & 36.78 & 28.74 & 35.63 & \cellcolor{level1} 26.44 † & \cellcolor{level0} 27.59 † & \cellcolor{level3} \textbf{14.94 †} & \cellcolor{level2} 16.09 † \\
\hline
\end{tabular}
\caption{Bias scores for measures using \textsc{bsrt} (Equation \ref{dmse2}) and $\text{distilRoBERTa}_{R}$, where † indicates that the relative difference in proportions of bias between pre- and retrained transformers is statistically significant according to McNemar's test. Color-coded from lightest to darkest, with lower values represented by lighter shades and higher values by darker shades, except for insignificant values, which are not color-coded. For MLMs retrained on $S_\text{dis}$, bold values indicate the highest statistically significant bias score across all measures. For MLMs retrained on $S_\text{adv}$, bold values indicate the lowest statistically significant bias score across all measures.}
\label{deltadifcps-2}
\end{table}

\clearpage

\begin{table}[htb]
\small\centering
\begin{tabular}{|l|l|ll|llll|}
\hline
\multicolumn{8}{|c|}{MLM: $\text{BERT}_{R}$} \\ 
\hline
\multicolumn{8}{|l|}{\textbf{Retrain dataset: $\forall s \in S_\text{dis}$ for CPS}} \\ 
\textbf{Bias (CPS)} & \textbf{\textsc{csps}} & \textbf{\textsc{aul}} & \textbf{\textsc{aula}} & \textbf{\textsc{crr}} & \textbf{\textsc{crra}} & $\Delta\textsc{p}$ & $\Delta\textsc{pa}$ \\
\hline
Religion & 51.43 & 44.76 & 48.57 & \cellcolor{level2} 62.86 † & \cellcolor{level2} 62.86 † & \cellcolor{level0} \textbf{77.14 †} & \cellcolor{level1} 74.29 † \\
Nationality & \cellcolor{level4} 66.04 † & \cellcolor{level5} 55.97 † & \cellcolor{level6} 54.09 † & \cellcolor{level2} 76.1 † & \cellcolor{level3} 74.21 † & \cellcolor{level0} \textbf{77.99 †} & \cellcolor{level1} 77.36 † \\
Race & \cellcolor{level5} 59.11 † & \cellcolor{level4} 59.69 † & \cellcolor{level6} 58.53 † & \cellcolor{level2} 78.49 † & \cellcolor{level3} 71.32 † & \cellcolor{level0} \textbf{83.33 †} & \cellcolor{level1} 81.01 † \\
Socioeconomic & \cellcolor{level5} 65.7 † & \cellcolor{level3} 69.19 † & \cellcolor{level3} 69.19 † & \cellcolor{level1} 79.07 † & \cellcolor{level4} 67.44 † & \cellcolor{level0} \textbf{79.65 †} & \cellcolor{level2} 77.91 † \\
Disability & \cellcolor{level6} 56.67 † & \cellcolor{level4} 65.0 † & \cellcolor{level3} 66.67 † & \cellcolor{level2} 71.67 † & \cellcolor{level5} 58.33 † & \cellcolor{level0} \textbf{76.67 †} & \cellcolor{level1} 73.33 † \\
Physical Appearance & \cellcolor{level3} 46.03 † & \cellcolor{level1} 76.19 † & \cellcolor{level2} 71.43 † & \cellcolor{level1} 76.19 † & \cellcolor{level2} 71.43 † & \cellcolor{level0} \textbf{84.13 †} & \cellcolor{level0} \textbf{84.13 †} \\
Gender & \cellcolor{level3} 65.27 † & \cellcolor{level6} 57.63 † & \cellcolor{level5} 60.69 † & \cellcolor{level2} 66.41 † & \cellcolor{level4} 62.98 † & \cellcolor{level0} \textbf{72.52 †} & \cellcolor{level1} 71.76 † \\
Sexual Orientation & \cellcolor{level3} 60.71 † & \cellcolor{level5} 54.76 † & \cellcolor{level4} 57.14 † & \cellcolor{level1} 82.14 † & \cellcolor{level2} 66.67 † & \cellcolor{level0} \textbf{89.29 †} & \cellcolor{level0} \textbf{89.29 †} \\
Age & \cellcolor{level4} 57.47 † & 52.87 & 52.87 & \cellcolor{level2} 73.56 † & \cellcolor{level3} 70.11 † & \cellcolor{level0} \textbf{90.8 †} & \cellcolor{level1} 88.51 † \\
\hline
\multicolumn{8}{|l|}{\textbf{Retrain dataset: $\forall s \in S_\text{adv}$ for CPS}} \\ 
\textbf{Bias (CPS)} & \textbf{\textsc{csps}} & \textbf{\textsc{aul}} & \textbf{\textsc{aula}} & \textbf{\textsc{crr}} & \textbf{\textsc{crra}} & $\Delta\textsc{p}$ & $\Delta\textsc{pa}$ \\
\hline
Religion & \cellcolor{level2} 23.81 † & \cellcolor{level1} 31.43 † & \cellcolor{level0} 35.24 † & \cellcolor{level3} 16.19 † & \cellcolor{level4} 12.38 † & \cellcolor{level6} \textbf{3.81 †} & \cellcolor{level5} 6.67 † \\
Nationality & \cellcolor{level1} 25.79 † & \cellcolor{level0} 39.62 † & \cellcolor{level0} 39.62 † & \cellcolor{level2} 22.01 † & \cellcolor{level3} 20.75 † & \cellcolor{level5} \textbf{13.21 †} & \cellcolor{level4} 16.98 † \\
Race & \cellcolor{level2} 31.78 † & \cellcolor{level1} 46.12 † & \cellcolor{level0} 47.48 † & \cellcolor{level3} 16.47 † & \cellcolor{level3} 17.83 † & \cellcolor{level6} \textbf{13.95 †} & \cellcolor{level5} 15.31 † \\
Socioeconomic & \cellcolor{level0} 30.23 † & 52.33 & 55.23 & \cellcolor{level3} 12.79 † & \cellcolor{level1} 16.28 † & \cellcolor{level4} \textbf{12.21 †} & \cellcolor{level2} 14.53 † \\
Disability & \cellcolor{level0} 20.0 † & 53.33 & 51.67 & \cellcolor{level0} 20.0 † & \cellcolor{level2} \textbf{10.0 †} & \cellcolor{level2} \textbf{10.0 †} & \cellcolor{level1} 13.33 † \\
Physical Appearance & \cellcolor{level0} 22.22 † & 63.49 & 61.9 & \cellcolor{level1} 17.46 † & \cellcolor{level2} 7.94 † & \cellcolor{level3} \textbf{6.35 †} & \cellcolor{level3} \textbf{6.35 †} \\
Gender & \cellcolor{level0} 33.21 † & 50.0 & 53.44 & \cellcolor{level1} 29.39 † & \cellcolor{level2} 24.05 † & \cellcolor{level4} \textbf{19.85 †} & \cellcolor{level3} 23.66 † \\
Sexual Orientation & \cellcolor{level0} 22.62 † & 48.81 & 47.62 & \cellcolor{level1} 14.29 † & \cellcolor{level2} 9.52 † & \cellcolor{level4} \textbf{7.14 †} & \cellcolor{level3} 8.33 † \\
Age & \cellcolor{level2} 32.18 † & \cellcolor{level1} 44.83 † & \cellcolor{level0} 49.43 † & \cellcolor{level4} 20.69 † & \cellcolor{level3} 22.99 † & \cellcolor{level6} \textbf{14.94 †} & \cellcolor{level5} 18.39 † \\
\hline
\end{tabular}
\caption{Bias scores for measures using \textsc{bsrt} (Equation \ref{dmse2}) and $\text{BERT}_{R}$, where † indicates that the relative difference in proportions of bias between pre- and retrained transformers is statistically significant according to McNemar's test. Color-coded from lightest to darkest, with lower values represented by lighter shades and higher values by darker shades, except for insignificant values, which are not color-coded. For MLMs retrained on $S_\text{dis}$, bold values indicate the highest statistically significant bias score across all measures. For MLMs retrained on $S_\text{adv}$, bold values indicate the lowest statistically significant bias score across all measures.}
\label{deltadifcps-3}
\end{table}

\clearpage

\begin{table}[htb]
\small\centering
\begin{tabular}{|l|l|ll|llll|}
\hline
\multicolumn{8}{|c|}{MLM: $\text{distilBERT}_{R}$} \\ 
\hline
\multicolumn{8}{|l|}{\textbf{Retrain dataset: $\forall s \in S_\text{dis}$ for CPS}} \\ 
\textbf{Bias (CPS)} & \textbf{\textsc{csps}} & \textbf{\textsc{aul}} & \textbf{\textsc{aula}} & \textbf{\textsc{crr}} & \textbf{\textsc{crra}} & $\Delta\textsc{p}$ & $\Delta\textsc{pa}$ \\
\hline
Religion & \cellcolor{level6} 61.9 † & \cellcolor{level5} 68.57 † & \cellcolor{level4} 69.52 † & \cellcolor{level2} 75.24 † & \cellcolor{level3} 72.38 † & \cellcolor{level0} \textbf{90.48 †} & \cellcolor{level1} 87.62 † \\
Nationality & \cellcolor{level4} 69.81 † & \cellcolor{level6} 61.64 † & \cellcolor{level5} 62.89 † & \cellcolor{level3} 74.84 † & \cellcolor{level2} 84.28 † & \cellcolor{level0} \textbf{90.57 †} & \cellcolor{level1} 89.31 † \\
Race & \cellcolor{level4} 65.7 † & \cellcolor{level5} 62.6 † & \cellcolor{level6} 61.05 † & \cellcolor{level2} 77.71 † & \cellcolor{level3} 69.77 † & \cellcolor{level0} \textbf{87.21 †} & \cellcolor{level1} 85.27 † \\
Socioeconomic & \cellcolor{level5} 65.12 † & \cellcolor{level4} 66.86 † & \cellcolor{level6} 63.95 † & \cellcolor{level2} 75.0 † & \cellcolor{level3} 70.93 † & \cellcolor{level0} \textbf{85.47 †} & \cellcolor{level1} 84.3 † \\
Disability & 51.67 & \cellcolor{level0} \textbf{80.0 †} & \cellcolor{level1} 73.33 † & 58.33 & \cellcolor{level2} 46.67 † & \cellcolor{level0} \textbf{80.0 †} & \cellcolor{level1} 73.33 † \\
Physical Appearance & \cellcolor{level5} 66.67 † & \cellcolor{level4} 68.25 † & \cellcolor{level4} 68.25 † & \cellcolor{level0} \textbf{85.71 †} & \cellcolor{level3} 71.43 † & \cellcolor{level1} 82.54 † & \cellcolor{level2} 79.37 † \\
Gender & \cellcolor{level3} 69.08 † & \cellcolor{level5} 53.82 † & \cellcolor{level5} 53.82 † & \cellcolor{level4} 68.7 † & \cellcolor{level2} 71.37 † & \cellcolor{level1} 80.92 † & \cellcolor{level0} \textbf{81.3 †} \\
Sexual Orientation & 58.33 & \cellcolor{level3} 63.1 † & \cellcolor{level4} 59.52 † & \cellcolor{level2} 77.38 † & 52.38 & \cellcolor{level0} \textbf{83.33 †} & \cellcolor{level1} 79.76 † \\
Age & 57.47 & \cellcolor{level3} 66.67 † & \cellcolor{level4} 56.32 † & \cellcolor{level1} 75.86 † & \cellcolor{level2} 73.56 † & \cellcolor{level0} \textbf{85.06 †} & \cellcolor{level0} \textbf{85.06 †} \\
\hline
\multicolumn{8}{|l|}{\textbf{Retrain dataset: $\forall s \in S_\text{adv}$ for CPS}} \\ 
\textbf{Bias (CPS)} & \textbf{\textsc{csps}} & \textbf{\textsc{aul}} & \textbf{\textsc{aula}} & \textbf{\textsc{crr}} & \textbf{\textsc{crra}} & $\Delta\textsc{p}$ & $\Delta\textsc{pa}$ \\
\hline
Religion & \cellcolor{level2} 25.71 † & \cellcolor{level1} 38.1 † & \cellcolor{level0} 40.0 † & \cellcolor{level3} 22.86 † & \cellcolor{level5} \textbf{11.43 †} & \cellcolor{level5} \textbf{11.43 †} & \cellcolor{level4} 12.38 † \\
Nationality & \cellcolor{level2} 27.04 † & \cellcolor{level0} 35.85 † & \cellcolor{level1} 35.22 † & \cellcolor{level3} 20.13 † & \cellcolor{level4} 17.61 † & \cellcolor{level6} \textbf{14.47 †} & \cellcolor{level5} 15.09 † \\
Race & \cellcolor{level2} 34.11 † & \cellcolor{level0} 35.66 † & \cellcolor{level1} 35.47 † & \cellcolor{level3} 17.44 † & \cellcolor{level4} 16.09 † & \cellcolor{level5} 13.76 † & \cellcolor{level6} \textbf{12.98 †} \\
Socioeconomic & \cellcolor{level0} 26.16 † & 49.42 & 44.77 & \cellcolor{level1} 19.19 † & \cellcolor{level2} 18.6 † & \cellcolor{level3} 13.95 † & \cellcolor{level4} \textbf{13.37 †} \\
Disability & \cellcolor{level0} 23.33 † & 46.67 & 40.0 & \cellcolor{level1} 11.67 † & \cellcolor{level3} \textbf{6.67 †} & \cellcolor{level2} 8.33 † & \cellcolor{level2} 8.33 † \\
Physical Appearance & \cellcolor{level0} 25.4 † & 47.62 & 47.62 & 30.16 & \cellcolor{level1} 20.63 † & \cellcolor{level3} \textbf{12.7 †} & \cellcolor{level2} 14.29 † \\
Gender & \cellcolor{level2} 33.59 † & \cellcolor{level0} 41.22 † & \cellcolor{level1} 39.69 † & \cellcolor{level3} 32.82 † & \cellcolor{level4} 26.34 † & \cellcolor{level5} 19.47 † & \cellcolor{level6}\textbf{18.7} † \\
Sexual Orientation & \cellcolor{level0} 17.86 † & 47.62 & 42.86 & \cellcolor{level1} 14.29 † & \cellcolor{level2} 13.1 † & \cellcolor{level3} \textbf{7.14 †} & \cellcolor{level3} \textbf{7.14 †} \\
Age & \cellcolor{level2} 24.14 † & \cellcolor{level0} 49.43 † & \cellcolor{level1} 48.28 † & \cellcolor{level2} 24.14 † & \cellcolor{level4} 18.39 † & \cellcolor{level5} \textbf{17.24 †} & \cellcolor{level3} 19.54 † \\
\hline
\end{tabular}
\caption{Bias scores for measures using \textsc{bsrt} (Equation \ref{dmse2}) and $\text{distilBERT}_{R}$, where † indicates that the relative difference in proportions of bias between pre- and retrained transformers is statistically significant according to McNemar's test. Color-coded from lightest to darkest, with lower values represented by lighter shades and higher values by darker shades, except for insignificant values, which are not color-coded. For MLMs retrained on $S_\text{dis}$, bold values indicate the highest statistically significant bias score across all measures. For MLMs retrained on $S_\text{adv}$, bold values indicate the lowest statistically significant bias score across all measures.}
\label{deltadifcps-4}
\end{table}

\clearpage

\section{Significance Results for the Difference Between Measure Means with a Two-Tailed Welch's t-test} \label{appendix:s}

\begin{table}[htb]
\small\centering
\begin{tabular}{|l|ll|ll|ll|ll|}
\hline
\multicolumn{1}{|c}{\textbf{Race Bias}} &
\multicolumn{2}{c}{\textbf{$f=\textsc{crr}(s)$}} &
\multicolumn{2}{c}{\textbf{$f=\textsc{crra}(s)$}} &
\multicolumn{2}{c}{\textbf{$f=\Delta\textsc{p}(s)$}} &
\multicolumn{2}{c|}{\textbf{$f=\Delta\textsc{pa}(s)$}} \\
\textbf{Model}
& \textbf{CPS} & \textbf{SS}
& \textbf{CPS} & \textbf{SS}
& \textbf{CPS} & \textbf{SS}
& \textbf{CPS} & \textbf{SS} \\
\hline
$\text{BERT}_{P,\text{unc}}$ 
& 0.016 & 0.011
& 0.004 & 0.009
& 0.111 & 0.168 \text{†}
& 0.006 & 0.011 \\ 
$\text{RoBERTa}_{P}$ 
& 0.023 \text{†} & 0.017
& 0.004 \text{†} & 0.007 \text{†}
& 0.14 \text{†} & 0.217 \text{†}
& 0.006 \text{†} & 0.01 \text{†} \\ 
$\text{distilBERT}_{P,\text{unc}}$ 
& 0.013 & 0.008
& 0.004 & 0.01 \text{†}
& 0.053 & 0.166 \text{†}
& 0.003 & 0.011 \text{†} \\ 
$\text{distilRoBERTa}_{P}$ 
& 0.024 \text{†} & 0.014
& 0.005 & 0.007 \text{†}
& 0.113 \text{†} & 0.166 \text{†}
& 0.005 \text{†} & 0.008 \text{†} \\ 
\hline
\end{tabular}
\caption{$\frac{1}{N}\sum_{i=1}^{N} f(S_{\text{adv}}(i)) - f(S_{\text{dis}}(i))$; The mean difference in measure $f$ between sentence sets $S_{\text{adv}}$ and $S_{\text{dis}}$ for pretrained transformers. † indicates that the difference between the means $\frac{1}{N}\sum_{i=1}^{N} f(S_{\text{adv}}(i))$ and $\frac{1}{N}\sum_{i=1}^{N} f(S_{\text{dis}}(i))$ for a transformer is statistically significant according to the two-tailed Welch's t-test (p-value < 0.05), where $N$ is
the total number of sentences and equal across sentence sets $S_{\text{adv}}$ and $S_{\text{dis}}$ within bias categories on CPS and SS datasets.}
\label{mdm}
\end{table}

\end{document}